\documentclass[pdflatex,sn-mathphys-ay]{sn-jnl}%

\usepackage{graphicx}%
\usepackage{multirow}%
\usepackage{amsmath,amssymb,amsfonts}%
\usepackage{amsthm}%
\usepackage{mathrsfs}%
\usepackage[title]{appendix}%
\usepackage{xcolor}%
\usepackage{textcomp}%
\usepackage{manyfoot}%
\usepackage{booktabs}%
\usepackage{algorithm}%
\usepackage{algpseudocode}%
\usepackage{listings}%
\usepackage{lmodern}
\usepackage{bm}
\usepackage{placeins}
\usepackage{anyfontsize}

\DeclareMathOperator*{\argmax}{arg\,max}

\theoremstyle{definition}
\newtheorem{prob}{Problem}
\newtheorem{definition}{Definition}
\newtheorem{theorem}{Theorem}
\newtheorem{lemma}{Lemma}
\begin{document}

\title[Summarising local explanations via proxies]{{\sc ExplainReduce}: generating global explanations from many local explanations%
}

\author*[1]{\fnm{Lauri} \sur{Sepp\"al\"ainen}}\email{lauri.seppalainen@helsinki.fi}

\author[1]{\fnm{Mudong} \sur{Guo}}\email{1198033143@qq.com}

\author[1]{\fnm{Kai} \sur{Puolam\"aki}}\email{kai.puolamaki@helsinki.fi}

\affil[1]{%
\orgname{University of Helsinki}, \orgaddress{\street{P.O. Box 64}, \city{Helsinki}, \postcode{00014}, \country{Finland}}}

\abstract{Most commonly used non-linear machine learning methods are closed-box models, uninterpretable to humans.
The field of explainable artificial intelligence (XAI) aims to develop tools to examine the inner workings of these closed boxes.
An often-used model-agnostic approach to XAI involves using simple models as local approximations to produce so-called local explanations; examples of this approach include {\sc lime},  {\sc shap}, and {\sc slisemap}.
This paper shows how a large set of local explanations can be reduced to a small ``proxy set'' of simple models, which can act as a generative global explanation.
This reduction procedure, {\sc ExplainReduce}, can be formulated as an optimisation problem and approximated efficiently using greedy heuristics.
We show that, for many problems, as few as five explanations can faithfully emulate the closed-box model and that our reduction procedure is competitive with other model aggregation methods.}


\maketitle

\section{Introduction}\label{sec:intro}
Explainable artificial intelligence (XAI) aims to elucidate the inner workings of ``closed-box'' machine learning (ML) models: models that are not readily interpretable to humans.
As machine learning has found applications in almost all fields, the need for interpretability has likewise led to the use of XAI in medicine \citep{band2023medical}, manufacturing \citep{peng2022industrial} and atmospheric chemistry \citep{seppalainen2023using}, among many other domains.
In the past two decades, many different XAI methods have been developed to meet the requirements \citep{guidotti2018survey, dwivedi2023survey} in this diverse set of fields.
XAI methods typically produce \emph{explanations}, i.e., distillations of a closed-box model's decision patterns.
These explanations can be divided into two categories by their applicability in the data space (local vs. global) and by their applicability to different ML models (model-specific vs. model agnostic).
Global explanations approximate the general behaviour of a closed-box model everywhere.
However, producing such explanations, especially of the model-agnostic variety, is often a challenge: if a model approximates a complex function, describing its behaviour may require describing the complex function itself, thereby defeating the purpose of interpretability.
Hence, a common approach to producing model-agnostic explanations is to relax the requirement for explaining the model globally and instead focus on local behaviour \citep{guidotti2018survey, mersha2024survey}.
Assuming a degree of smoothness, approximating the closed-box function in a small neighbourhood is often feasible using simple, interpretable functions, such as sparse linear models, decision trees, or decision rules.

This work was motivated by our observation that in practice, the local explanations for similar items will often be similar, provided the underlying closed-box function is smooth in the region.
Hence, many local explanations may explain a given item with nearly equal fidelity.
We initially observed this phenomenon with {\sc slisemap}, which embeds the datapoints and their embeddings in a plane based on the similarity of their local explanations.
We found that most items could be approximated with almost equal fidelity by using any of the neighbouring local models; see Sect. 4.5 in \citet{bjorklund2023slisemap}.
Conversely, many local models can accurately approximate multiple items.
These two observations suggest that many local explanations are redundant.
Furthermore, if we generate a large set of local explanations, we may be able to find a smaller subset thereof, which could effectively replace the full set without sacrificing the fidelity of the explanations.

\begin{figure}
    \centering
    \includegraphics[width=\textwidth]{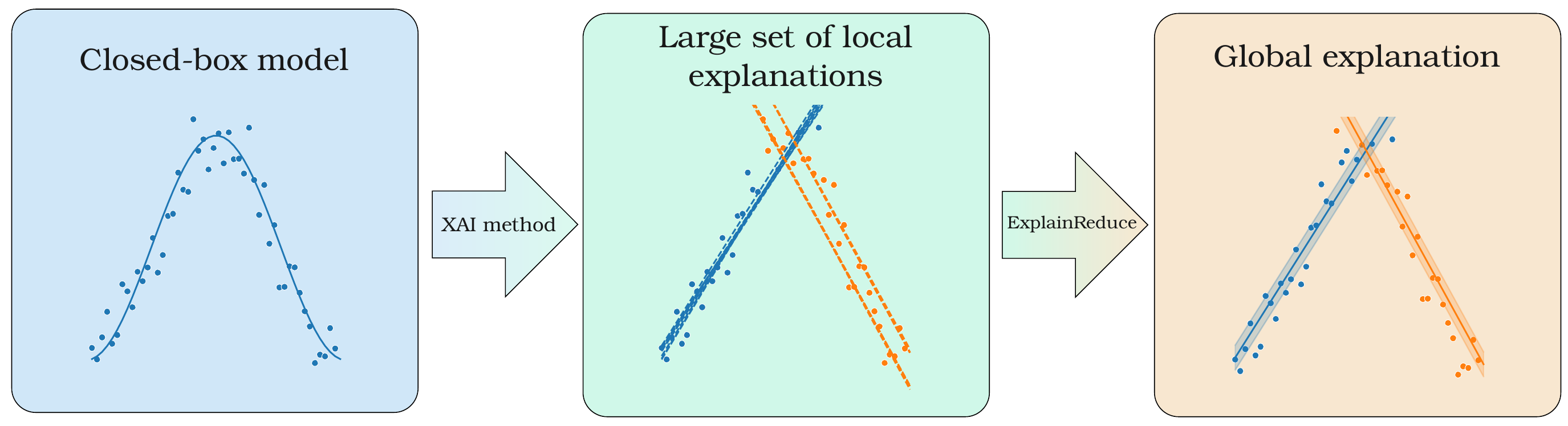}
    \caption{A schematic of the {\sc ExplainReduce} procedure. A closed-box model (left) can have many local explanations (middle). We can reduce the size of the local explanation set to get a global explanation consisting of two simple models (right).}
    \label{fig:pyramid_example}
\end{figure}

In this paper, we introduce a procedure, coined {\sc ExplainReduce}, that can reduce large sets of local explanations to a small subset of so-called proxy models.
This small proxy set can be used as a global explanation for the closed-box model, among other possible applications.
Fig. \ref{fig:pyramid_example} provides insight into the process.
In the left panel, we show noisy samples (blue dots) from a closed-box function (marked with a solid blue line).
The middle panel shows a set of local explanations (here linear models,  blue and orange dashed lines) produced by the XAI method {\sc slisemap} \citep{bjorklund2023slisemap}, with one explanation per data item.
The right panel shows a reduction from this large set of local models to two (solid lines), maximising the coverage of items within an error tolerance (shaded area).
The procedure offers a trade-off between interpretability, coverage, and fidelity of the local explanation model by finding a minimal subset of local models that can still approximate the closed-box model for most points with reasonable accuracy, as demonstrated in Section \ref{sec:results}.
Our approach differs from prior explanation aggregation methods in three ways: the procedure (i) can be applied to both regression and classification tasks, (ii) can be applied to many types of local explanations, and (iii) uses efficient approximate optimisation approaches with an upper bound on worst-case performance (see Section \ref{sec:methods}).

The contributions of this work are as follows:
\begin{enumerate}
    \item We introduce ExplainReduce, a model-agnostic procedure for reducing large sets of local explanations into compact proxy sets that serve as global explanations.
    \item We formalize explanation reduction as an optimization problem balancing fidelity, coverage, and interpretability, and propose efficient greedy algorithms with theoretical guarantees (Section \ref{sec:methods}).
    \item We empirically demonstrate that hundreds of local post-hoc explanations can be reduced to a representative subset (e.g., five models) while retaining the fidelity of the full explanation set (Section \ref{sec:results}).
    \item We compare {\sc ExplainReduce} against existing aggregation methods, showing superior performance and scalability (Section \ref{ssec:global_comp}).
\end{enumerate}
The code for the procedure and to recreate each of the experiments presented in this paper can be found in the supplementary material.

\section{Applications for reduced local explanation sets}

Here, we list some use cases for our proposed algorithm.

\textbf{Global explanations}:
Generating one local explanation per data point easily results in hundreds, if not thousands, of explanations, overwhelming users.
A reduced proxy set offers a compact, interpretable global view of the closed-box model behaviour.

\textbf{Interpretable surrogate for the closed-box model}:
If local explanations faithfully approximate the closed-box model, the full set of these explanations can replace it with minimal accuracy loss.
A small proxy set offers a practical, interpretable surrogate without sacrificing accuracy.

\textbf{Exploratory data analysis and XAI method comparison}:
Studying reduced explanation sets can reveal redundancy in explanations and highlight clusters of similar models.
Comparing proxy sets from different XAI methods offers insights into the stability and consistency of their explanations.

\textbf{Outlier detection}:
Given a novel observation, we can assess how well the local models in the reduced set predict it relative to similar items in the training dataset.
If the novel item is better explained by a different model than the one used for other similar items, or if no explanation in the reduced set accurately captures the relationship, the novel item could be considered an outlier.

\textbf{Clustering}:
By associating each training item with an explanation from a small set, we can partition the data based on the closed-box model behaviour.
If the association is achieved, for example, by selecting the explanation with the lowest error, we can perform clustering without a distance measure, which is a desirable property for many types of data for which constructing a distance measure is non-trivial, such as images or molecules.

\section{Related work}\label{sec:related}

Modern machine learning tools have become indispensable in almost all industries and research fields \citep{rashid2024aisurvey}.
As these tools find more and more applications, awareness of their limitations has simultaneously spread.
Chief among these is the lack of interpretability inherent to many of the most potent ML methods \citep{mersha2024survey, ali2023xai}.
Understanding why ML models produce a specific prediction instead of something else poses obstacles in their adoption in, e.g., high-trust applications \citep{dosilovic2018survey}.
Such opaque methods are colloquially called black- or closed-box methods, in contrast with white- or open-box methods, characterised by their interpretability to humans.
Examples of closed-box methods include random forests and deep neural networks, whereas, e.g., statistical models like linear regression models are often considered open-box methods \citep{molnar:2019:a}.
As a result, the field of XAI has grown substantially in the last decade \citep{mersha2024survey}.

XAI methods generally take a dataset and a closed-box function as inputs and produce an explanation that describes the relationship between the inputs and outputs of the closed-box model.
One commonly used approach is to produce post-hoc explanations using local (open-box) surrogate models \citep{burkhart2021survey}.
In this approach, given a closed-box method $f(\bm{X}) = \hat{\bm{y}}$, a local surrogate model $g$ is another function that replicates the behaviour of the closed-box method in some region of the input space.

Many methods have been proposed to produce post-hoc local surrogate models as explanations.
In this work, we focus mainly on local linear approximations, generated via three general methods.
The first category uses gradient information either directly, termed {\sc vanillagrad} in this article, or by smoothing via sampling the neighbourhood ({\sc smoothgrad} by \citet{smilkov2017smoothgrad}).
The next category is using the closed-box model as an oracle for items sampled from the neighbourhood of the item under explanation.
This category contains {\sc lime} \citep{ribeiro2016} and {\sc shap} \citep{lundberg2017unified}.
Finally, we consider {\sc slisemap} \citep{bjorklund2023slisemap} and its variant {\sc slipmap} \citep{bjorklund2024SLIPMAP}, which avoid sampling by simultaneously finding a low-dimensional embedding for the local explanations alongside the explanations themselves.
In addition to local linear surrogates, we also consider rule-based explanations, where local explanations take the form of simple decision rules.
We use {\sc lore} \citep{guidotti2019lore} as the exemplary method.
For a more detailed description of the XAI methods mentioned above, we refer to Appendix \ref{a:xai}.

A shared property of post-hoc local surrogate models is the lack of uniqueness; for a given item, many local surrogates may exist with similar performance.
The phenomenon is documented for {\sc slisemap} in \citep{bjorklund2023slisemap} and implied for other methods based on the connection between loss landscapes of neural networks and the instability of explanations \citep{dombrowski2019explanations}, as well as for many publications aimed at fixing the inherent instability of {\sc lime} \citep{shankar2019alime, zafar2019dlime, zhao2021baylime}.

Several methods have been proposed to summarise local explanations into a representative set.
In their paper describing {\sc lime} \citep{ribeiro2016}, the authors propose an approach called \emph{submodular pick}, which balances global feature importance scores with model diversity.
This limits the method's applicability to feature-attribution-based explanations.
{\sc GLocalX} \citep{setzu2021glocalx} is a method of merging rule-based explanations to find a global explanation.
Consequently, the method is limited in applicability to rule-based explanations for classification models.
\cite{li2022optimal} proposes an integer-programming-based approach to finding an optimal set of local explanations.
In this approach, the authors aim to maximise the coverage (defined as the number of items in the training set within a radius $r$ of the nearest explanation in the reduced set) while minimising the 0--1 loss of the reduced explanation set.
Their approach is also limited to classification only and hinges on a well-defined distance measure in the data space.
Furthermore, the complicated IP problem for finding a small explanation set is computationally expensive to solve.

A related paradigm is the so-called case-based or example-based approaches, which summarise data rather than explanations and use representative samples in the training set to explain novel ones \citep{agnar1994casebased, kim2016examples}.
One such method is using prototypes, which present the user with the most similar ``prototype items'' as an explanation, such as showing images of birds with similar plumage as a basis for classification. 

{\sc ExplainReduce} expands the prior work in three key aspects.
The method is not limited to either regression or classification problems. It can be applied to a wide set of local XAI methods, provided that the explanation can be used as a surrogate model, i.e., can produce predictions for previously unseen points.
Furthermore, we formulate explanation reduction as a set-cover-like optimisation problem with guaranteed worst-case behaviour, and show that our method can find the reduced explanation sets very efficiently using greedy approximations.

\section{Methods}\label{sec:methods}

{\sc ExplainReduce} aims to summarise a large set of local explanations into a small, representative subset -- called \emph{proxy models} -- that collectively approximates the behaviour of the original closed-box model.
This approach combines ideas from local surrogate explanations and prototype selections: instead of finding representative data items, we select representative explanations.

The procedure consists of three steps: (i) generate a large number of local explanations using any pre-existing XAI method, (ii) formalise the reduction task as an optimisation problem, and (iii)
solve the problem using greedy algorithms with theoretical worst-case guarantees.

Below, we define the reduction problem and describe algorithms and evaluation metrics.

\subsection{Problem definition}
\label{sec:problem}
A dataset $\mathcal{D} = \{(\bm{x}_1, \bm{y}_1) , \ldots, (\bm{x}_n, \bm{y}_n)\}$ consists $n$ of data items (covariates) $\bm{x}_i \in \mathcal{X}$ and labels (responses) $\bm{y}_i \in \mathcal{Y}$. We use ${\bm X}$ to denote a matrix of $n$ rows such that ${\bm X}_{i\cdot}=\bm{x}_i$ when ${\cal X}$ is a real vector space (as in the examples of this paper).
A \emph{closed-box model}, i.e., a trained supervised learning algorithm $f(\bm{x}_i) = \hat{\bm{y}}_i$, produces predictions from items to labels.
It should be noted that much of the analysis in this section applies even if an explicit closed-box model is not provided.
In such cases, we treat the data-generating process as the closed-box model ($\bm{\hat{y}}_i = \bm{y}_i$).

A \emph{local explanation} for a data item $(\bm{x}_i, \bm{\hat{y}}_i)$ is a simple model $g(\bm{x}_i) = \Tilde{\bm{y}}_i$  which locally approximates either the connection between the data items and the labels or the behaviour of the closed-box functions.
Let $\bm{G} = (g_1, \dots, g_m)$ be a set of $m$ such explanations, obtained, e.g., by an XAI model providing local explanations.
Next, we define a loss function $\ell:\mathcal{Y} \times \mathcal{Y} \rightarrow \mathbb{R}_{\geq 0}$ and a loss matrix $\bm{L}\in{\mathbb{R}}_{\ge 0}^{m\times n}$, where individual items are defined as $\bm{L}_{ij} = \ell(g_i(\bm{x}_j), \bm{\hat{y}}_j)$.
For a given $\varepsilon\in{\mathbb{R}}_{>0}$, we say an explanation $g_i$ is \textit{faithful} for a data item $j$ if $\ell (g_j(\bm{x}_i), \bm{\hat{y}}_i) \leq \varepsilon$.
Finally, assume that the set of explanations $\bm{G}$ contains at least one faithful approximation for each item in the dataset ${\cal D}$. %
This can be achieved by, e.g., learning a local explanation for each item in ${\cal D}$.

We are trying to find an optimal subset of models $\bm{S} \subseteq[m]=\{1,\ldots,m\}$, which we call a \emph{proxy set}. 
We want the proxy set $\bm{S}$ to be faithful to the closed-box model. %
Additionally, we are interested in how many items can be explained by a given set of local surrogates to a satisfactory degree.
To measure this, we use coverage $C$, defined as the proportion of data items which can be explained sufficiently by at least one local model in the subset $\bm{S}$.
Mathematically, given a loss threshold $\varepsilon\in{\mathbb{R}}_{>0}$, the coverage can be calculated as
\begin{equation}\label{eq:C}
    C({\bm{S}}, \varepsilon) = \left| \{j\in [n]\mid\min\nolimits_{i\in{\bm{S}}}{} \ell(g_i(\bm{x}_j), \bm{\hat{y}}_j) \leq \varepsilon \; \} \right|/n.
\end{equation}
A $c$-covering subset of local models ${\bm{S}}$ is a set for which $C({\bm{S}}, \varepsilon) \geq c$.

Next, we define three computational problems to address the task described earlier, alongside the algorithms we use to solve these problems.

\subsubsection{Maximum coverage.} The first formulation attempts to minimise the number of items for which a satisfactory local explanation is not included in the subset ${\bm{S}}_c$.
Conceptually, we aim to maximise the probability that for a given item, the proxy set $\bm{S}$ contains at least one local model such that the loss for that item is within the set error tolerance.
\begin{prob} ({\sc Max Coverage})\label{prob:1}
Given $k$ and $\varepsilon\in{\mathbb{R}}_{>0}$, find a subset ${\bm{S}}_c$ of cardinality $k$ that maximises coverage, or
\begin{equation}\label{eq:p1}
{\bm{S}}_c^k=\arg\max\nolimits_{{\bm{S}}\subseteq [m]_k} C({\bm{S}}, \varepsilon), 
\end{equation}
where we have used $[m]_k=\{{\bm S}\subseteq[m]\mid\left|{\bm S}\right|=k\}$ to denote the subsets of cardinality $k$.
\end{prob}
Problem~\ref{prob:1} is a variant of the NP-complete partial set covering problem, sometimes called the {\sc max k-cover}.
The partial set covering problem is an example of submodular maximisation.

We solve Problem~\ref{prob:1} using a standard iterative greedy algorithm, which we call {\sc reduce} (Alg. \ref{alg:reduce}), which takes as an input the cardinality $k$ and the marginal gain metric $\Delta(i \mid \bm{S})$ for which we use the marginal increase of coverage:
\begin{equation}\label{eq:deltaC}
\Delta C(i \mid \bm{S}) = C(\bm{S} \cup \{i\},\varepsilon) - C(\bm{S},\varepsilon).
\end{equation}

\begin{algorithm}
\caption{{\sc reduce}: Simple greedy optimisation.}
\label{alg:reduce}
\hspace*{\algorithmicindent} \textbf{Input:} $k$: target size; $\Delta(i \mid \bm{S})$: marginal gain function; initialise $\bm{S} \gets \emptyset$. \\
\hspace*{\algorithmicindent} \textbf{Output:} $\bm{S}\subseteq [m]$: subset of indices.
\begin{algorithmic}
    \Procedure{reduce}{ $k, \Delta$}
        \While{$|\bm{S}| < k$}
            \State $i^* \gets \argmax_{i \in [m] \setminus \bm{S}} \Delta(i \mid \bm{S})$ ; $\bm{S} \gets \bm{S} \cup \{i^*\}$
        \EndWhile\\
        \Return $\bm{S}$
    \EndProcedure
\end{algorithmic}
\end{algorithm}

The greedy algorithm has a lower bound of achieving coverage at least $1 - \left((k - 1)/k\right)^k$ ($\ge 1-1/e\approx 0.632$) times the optimal solution \cite{nemhauser1978analysis}\footnote{Full proof in Appendix \ref{a:proofs}}.
We refer to this algorithm with {\sc Max Coverage}.

\subsubsection{Minimum loss.} The second definition aims to capture a subset of explanations that can serve as a proxy model with a small average loss.
This is equivalent to maximising the faithfulness of the proxy set to the closed-box model.
\begin{prob} ({\sc Min Loss})\label{prob:2}
Given $k$, find a subset ${\bm{S}}_c$ with cardinality $k$ such that the average loss $\mathcal{L}(\bm{S})$ when picking the lowest loss model from ${\bm{S}}_c$ is minimised, or
\begin{align}
{\bm{S}}_c&=\arg\min\nolimits_{{\bm{S}}\in[m]_k} \mathcal{L}(\bm{S})\quad\text{where}\\
\mathcal{L}(\bm{S}) &= \left(\sum\nolimits_{j=1}^n{\min\nolimits_{i\in{\bm{S}}}{\ell(g_i(\bm{x}_j),\bm{\hat{y}}_j)}} /n\right). \label{eq:loss}
\end{align}
\end{prob}
While minimizing $\mathcal{L}(\bm{S})$ is a supermodular minimization problem, it can be framed as the maximization of a non-negative, monotone submodular function $f(\bm{S})=\mathcal{L}_{base}-\mathcal{L}(\bm{S})$, where $\mathcal{L}_{base}$ represents the average loss of an initial reference point (e.g., the loss of an empty set or a single starting model).
In this framework, $f(\bm{S})$ quantifies the total reduction in loss achieved by the proxy set.

By applying the greedy algorithm (\textsc{reduce}) to this improvement function, we leverage the classic result by \cite{nemhauser1978analysis}: the greedy solution $\bm{S}_g$ is guaranteed to capture at least $(1-1/e)\approx63.2\%$ of the total improvement possible by the optimal subset $\bm{S}^*$ of the same size.
Formally, $\mathcal{L}_{base}-\mathcal{L}(\bm{S}_g) \geq (1-1/e)(\mathcal{L}_{base}-\mathcal{L}(\bm{S}^*))$.
In practice, we can solve the problem using as marginal gain metric the marginal increase in negative loss:
\begin{equation}\label{eq:deltaL}
\Delta \mathcal{L}(i | \bm{S}) =  \mathcal{L}(\bm{S})-\mathcal{L}(\bm{S} \cup \{i\}) .
\end{equation}
We refer to this algorithm with {\sc Min Loss}.

\subsubsection{Joint utility maximisation.} 
While Problems \ref{prob:1} and \ref{prob:2} address coverage and faithfulness individually, a practical proxy set must balance both.
We propose a third formulation that treats the proxy selection task as multi-objective optimisation problem.
\begin{prob} ({\sc Balanced})\label{prob:3}
Given $k$ and $\varepsilon\in{\mathbb{R}}_{>0}$ and a tradeoff parameter $\lambda\in[0,1]$, find a subset $\bm{S}_c$ of cardinality $k$ such the joint utility of coverage and normalised loss reduction, $U(\bm{S})$, is maximised, or
\begin{align}
{\bm{S}}_c&=\arg\max\nolimits_{{\bm{S}}\in[m]_k} U(\bm{S})\quad\text{where}\\
U(\bm{S}) &= \lambda \cdot C(\bm{S}, \varepsilon) + (1 -\lambda)\cdot\left(\frac{\mathcal{L}_{base} - \mathcal{L}(\bm{S})}{\mathcal{L}_{base}} \right), \label{eq:utility}
\end{align}
where $\mathcal{L}_{base}$ is the initial average loss (e.g., the loss of the data-generating process or a single-model baseline).
\end{prob}

Since $U(\bm{S})$ is a linear combination of two monotone submodular functions, $U(\bm{S})$ is itself monotone submodular.
Consequently, solving Problem~\ref{prob:3} via Alg. \ref{alg:reduce} provides constant-factor approximation guarantee.
Specifically, the greedy solution $\bm{S}_g$ achieves utility such that $U(\bm{S}_g) \geq (1-1/e)U(\bm{S}^*)$, where $\bm{S}^*$ is the optimal subset of size $k$.
In this paper, we use $\lambda=0.5$ to balance coverage and fidelity; in our tests (Appendix \ref{a:cov_eps_sensitivity}), the results were not overly sensitive to $\lambda$, making $\lambda=0.5$ a straightforward choice. 
We refer to this algorithm with {\sc Balanced}.

While we have presented worst-case guarantees for our approaches to solving Problems~\ref{prob:1}-\ref{prob:3}, in our experiments, the performance of these greedy algorithms is close to the true optimum (Appendix \ref{a:greedy}).

\subsection{General procedure}\label{sec:procedure}

The {\sc ExplainReduce} algorithm is outlined in Algorithm \ref{alg:main}. 
The algorithm works in three stages.
First, a large set of local explanations $\bm{G}$ is generated using an XAI method on the dataset $\mathcal{D}$ and a closed-box model $f$.
Second, we use an optimisation algorithm {\sc reduce} to find an approximately optimal proxy set $\bm{S}$.
The options for the optimisation algorithm are described in Section \ref{sec:problem} above.
Finally, each data item in $\mathcal{D}$ is mapped to a proxy model in $S$ that best approximates the closed-box model for that item, i.e., has the lowest loss.
The algorithm outputs the optimal proxy set $\bm{S}_c$ and the aforementioned mapping.

\begin{algorithm}
\caption{{\sc ExplainReduce} Procedure to find a subset of explanations.}
\label{alg:main}
\hspace*{\algorithmicindent} \textbf{Input:} $\mathcal{D} \gets \{(\bm{x}_i, \hat{\bm{y}}_i) | i \in [n]\}$: dataset; ${\bm G}$: the set of $m$ local explanations; $\textrm{reduce}$: method to find ${\bm S}_c$, parametrised optionally by $\varepsilon$, $c$, or $k$, see Sect. \ref{sec:problem}. \\
\hspace*{\algorithmicindent} \textbf{Output:} ${\bm S}_c$: reduced set of explanations; $\text{map}$: $\text{map}[i]: [n]\mapsto{\bm S}_c$, mapping from the local dataset $[n]$ to explanations in ${\bm S}_c$.
\begin{algorithmic}
    \Procedure{ExplainReduce}{}    
    \State $\bm{S}_c \gets \textrm{reduce}(\bm{G}, \varepsilon, c, k)$ \Comment{$\textrm{reduce}$ is defined in Sect. \ref{sec:problem}}
    \State $\text{map} \gets \{\}$ \Comment{a mapping between items in $\mathcal{D}$ and the local models}
    \For{$i \in [n]$}
       \State $\text{map}[i] \gets \arg\min_{j\in{\bm S}_c}{\ell(g_j(\bm{x}_i), \bm{y}_i)}$
    \EndFor \\
    \Return ${\bm S}_c,\; \text{map}$\\
    \EndProcedure
\end{algorithmic}
\end{algorithm}

\subsection{Performance measures}

\bmhead{Fidelity} Fidelity \citep{guidotti2018survey} measures the adherence of a surrogate model to the complex closed-box function.
Given a closed-box function $f$, we fidelity is the loss between the closed-box model prediction $\hat{\bm{y}} = f(\bm{x})$ and the surrogate model prediction:
\begin{equation}
    \textrm{fidelity} =  \sum\nolimits_{i=1}^{n} \ell(g_i(\bm{x}_i), \hat{y}_i)/n,
    \label{eq:fidelity}
\end{equation}
where $g_i$ is the relevant local model, either from the full set of local explanations or the proxy model set.
Somewhat paradoxically, lower fidelity is better.
In this paper, we use mean square error as the loss function for regression tasks.
For classification tasks, we use the squared Hellinger distance of class probabilities,
\begin{equation}
    \ell_c(\hat{\bm{y}}, \tilde{\bm{y}}) = \sum\nolimits_{i=1}^{n} \left\|\sqrt{\hat{\bm{y}}_i} - \sqrt{\tilde{\bm{y}}_i}\right\|^2_2/2,
\end{equation}
where $\tilde{\bm{y}_i} = g_i(\bm{x}_i)\in\Delta^{p-1}$, where $\Delta^{p-1}$ is the p-simplex containing class probabilities of the $p$ classes; a more numerically stable replacement for the Kullback-Leibler divergence as in \citet{bjorklund2023slisemap}.%

\bmhead{Instability} Instability \citep{guidotti2018survey} (sometimes also referred to as \emph{stability}, despite lower values denoting better performance) measures how much a slight change in the input changes the explanation.
In practice, we model the slight change by measuring the loss of a given local model $g_i$ associated with item $\bm{x}_i$ with its $\kappa$ nearest neighbours:
\begin{equation}
    \textrm{instability} =  \sum\nolimits_{i=1}^{n} \sum\nolimits_{j \in \textrm{NN}_\kappa(i)} \ell(g_i(\bm{x}_j), \hat{y}_j)/(n\,\kappa). 
    \label{eq:instability}
\end{equation}
In this work, we use a fixed number of $\kappa=5$ nearest neighbours to report instability values ($\left|\textrm{NN}_5(i)\right|=5$).

\section{Results}
\label{sec:results}

In this section, we evaluate {\sc ExplainReduce} on a diverse set of regression and classification tasks and XAI methods to assess its ability to produce compact, interpretable global explanations without sacrificing fidelity.
Our experiments seek to address three key questions: (i) how well do proxy sets approximate the original closed-box model compared to the full set of local explanations, (ii) how does performance vary with the proxy set size and the number of initial local explanations, and (iii) how does {\sc ExplainReduce} compare to existing aggregation methods.

We run experiments on twelve datasets containing both synthetic and real-world examples (Table \ref{tab:datasets}, Appendix \ref{a:datasets}).
For each dataset, we train a closed-box model, generate local explanations using multiple XAI methods and apply {\sc ExplainReduce} to find proxy sets.
The closed-box models have been chosen based on task-specific performance.
In cases where multiple model types showed similar results, closed-box models were chosen to maximise model type diversity.
Performance is measured primarily on test data---items not used to generate the initial set of local explanations---to best capture general adherence to the closed-box model.
In this manner we can measure whether the explanation set contains a reasonable explanation for any item.

\begin{table}[b]
    \centering
    \resizebox{\textwidth}{!}{
    \begin{tabular}{l|cccc}
\hline
Dataset Name    & Size        & Task           & Closed-box model  & Citation                                     \\ \hline
Synthetic       & 5000 × 11   & Regression     & Random Forest     & \citep{bjorklund2023slisemap}                \\
Air Quality     & 7355 × 12   & Regression     & Random Forest     & \citep{oikarinenDetectingVirtualConcept2021} \\
Life Expectancy & 2938 × 22   & Regression     & Neural Network    & \citep{rajarshi2017life}                     \\
Vehicle         & 2059 × 12   & Regression     & SVR               & \citep{birla2022vehicles}                    \\
Gas Turbine     & 36733 × 9   & Regression     & Adaboost          & \citep{2019gas}                              \\
QM9             & 133766 × 27 & Regression     & Neural Network    & \citep{ramakrishnan2014Quantum}              \\
Adult           & 48842 × 14  & Classification & Neural Network    & \citep{becker1996adult}                      \\
Churn           & 5000 × 20   & Classification & Random Forest     & \citep{blake1998churn}                       \\
HIGGS           & 100000 × 28 & Classification & Gradient Boosting & \citep{whiteson2014higgs}                    \\
Jets            & 266421 × 7  & Classification & Random Forest     & \citep{CMS:opendata}                         \\
Spam            & 4601 × 57   & Classification & Gradient Boosting & \citep{hopkins1999spam}                      \\
Telescope       & 13376 × 10  & Classification & Neural Network    & \citep{bock2004telescope}                    \\ \hline
\end{tabular}}
    \caption{Summary of datasets used in experiments. Detailed dataset descriptions available in Appendix \ref{a:datasets}.}
    \label{tab:datasets}
\end{table}

A practical challenge is that local explanation sets do not have an obvious mapping from unseen items to explanations.
To address this, we adopt a simple nearest-neighbour approach: each novel item is assigned to a proxy corresponding to the training item that is closest in the feature space using Euclidean distance.
While using this approach has obvious limitations, we nonetheless found it to work reasonably well, as will be demonstrated.
In practical implementations, when using {\sc ExplainReduce} with novel items, we recommend that users use whatever distance measure is most fitting for their data.

The following subsections present case studies to build intuition, followed by systematic comparisons of reduction algorithms and baselines.

\subsection{Case studies}

\subsubsection{Synthetic data}

To give a simple example of the procedure, we apply it to a synthetic dataset.
The dataset, which is described in detail in Appendix \ref{a:datasets}, is generated by producing $k=4$ clusters in the input space and generating a random, different linear model for each cluster.
We then generate labels by applying a local model to the data items based on their cluster ID and adding Gaussian noise.
The dataset is then randomly split into a training set and a test set, and we train a {\sc smoothgrad} explainer on the training data.
For reduction, we use a greedy {\sc max coverage} algorithm, where $\varepsilon$ is defined as the 10th percentile of the loss matrix $\bm{L}$.

\begin{figure}
    \centering
    \includegraphics[width=\textwidth]{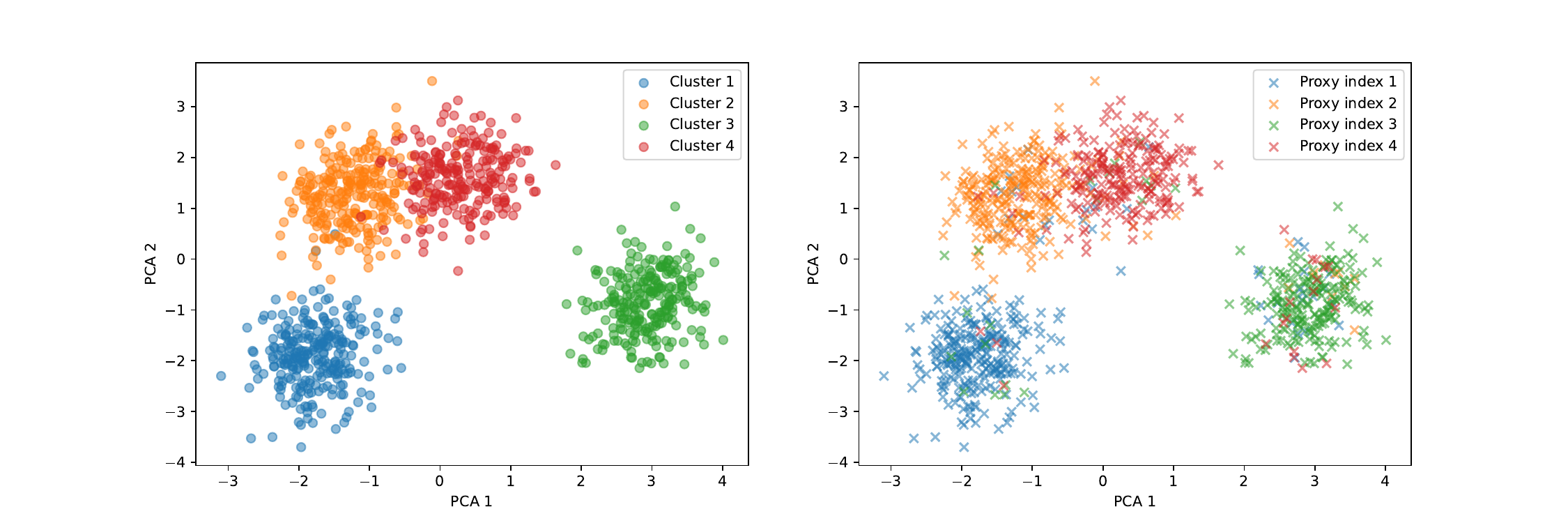}
    \caption{{\sc ExplainReduce} is able to correctly associate items with the correct local explanation. The left panel shows a PCA embedding of a synthetic test dataset with ground truth cluster labels, and the right panel shows the same dataset where colours correspond to reduced model indices. The XAI method used here is {\sc smoothgrad}, and the {\sc max coverage} algorithm was used for the reduction. We can see that the reduction faithfully approximates the ground truth clustering.}
    \label{fig:case_PCA}
\end{figure}

\begin{figure}
    \centering
    \includegraphics[width=\textwidth]{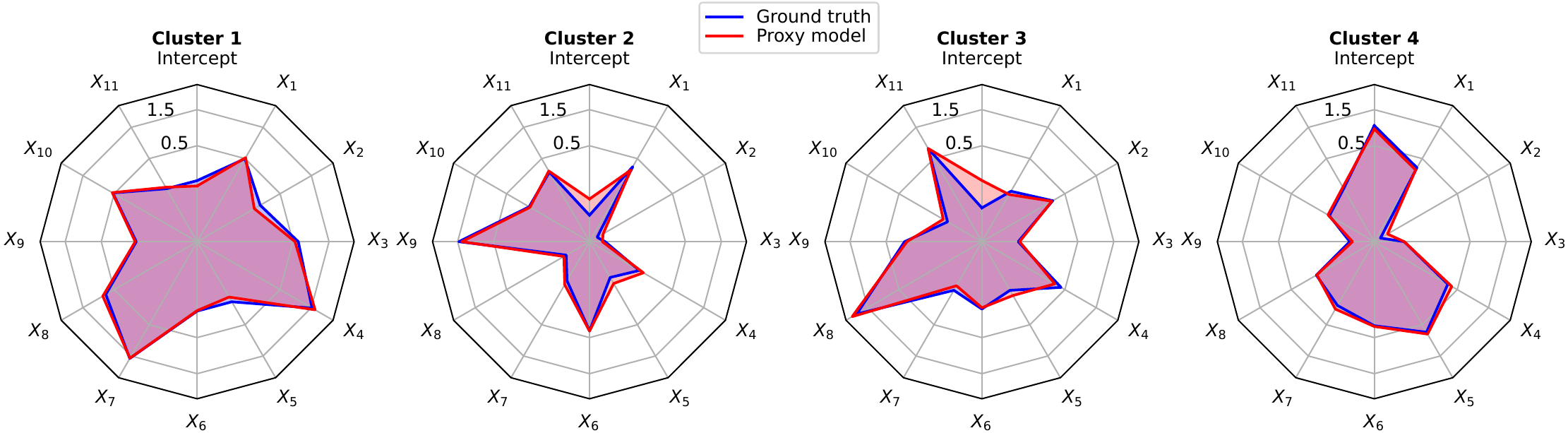}
    \caption{The proxy models found by {\sc ExplainReduce} (red) correspond well to ground truth local models (blue) used to generate a synthetic dataset. The spokes in the radar plots correspond to linear model coefficients associated with each feature.}
    \label{fig:case_radar}
\end{figure}

In Fig. \ref{fig:case_PCA}, we show a PCA of the items in the test set on the left, coloured with the ground truth cluster labels, and how the test items are mapped to proxy models on the right.
Overall, we can see that most items get mapped to the correct proxy model for that particular cluster.
The small impurity in the clusters stems from both the Gaussian noise and the approximative nature of the greedy coverage-maximising algorithm.
Furthermore, as Fig. \ref{fig:case_radar} shows, the reduced proxy models (red) correspond well with the ground truth models (blue).
The proxy set thus serves well as a generative global explanation for the dataset.
\subsubsection{Particle jet classification}

In a previous work \citep{seppalainen2023using}, we analysed a dataset containing simulated LHC proton-proton collisions using {\sc slisemap}.
These collisions can create either quarks or gluons, which decay into cascades of stable particles called jets.
These jets can then be detected, and we can train the classifier to distinguish between jets created by quarks and gluons based on the jets' properties.

\begin{figure}
    \centering
    \includegraphics[width=\textwidth]{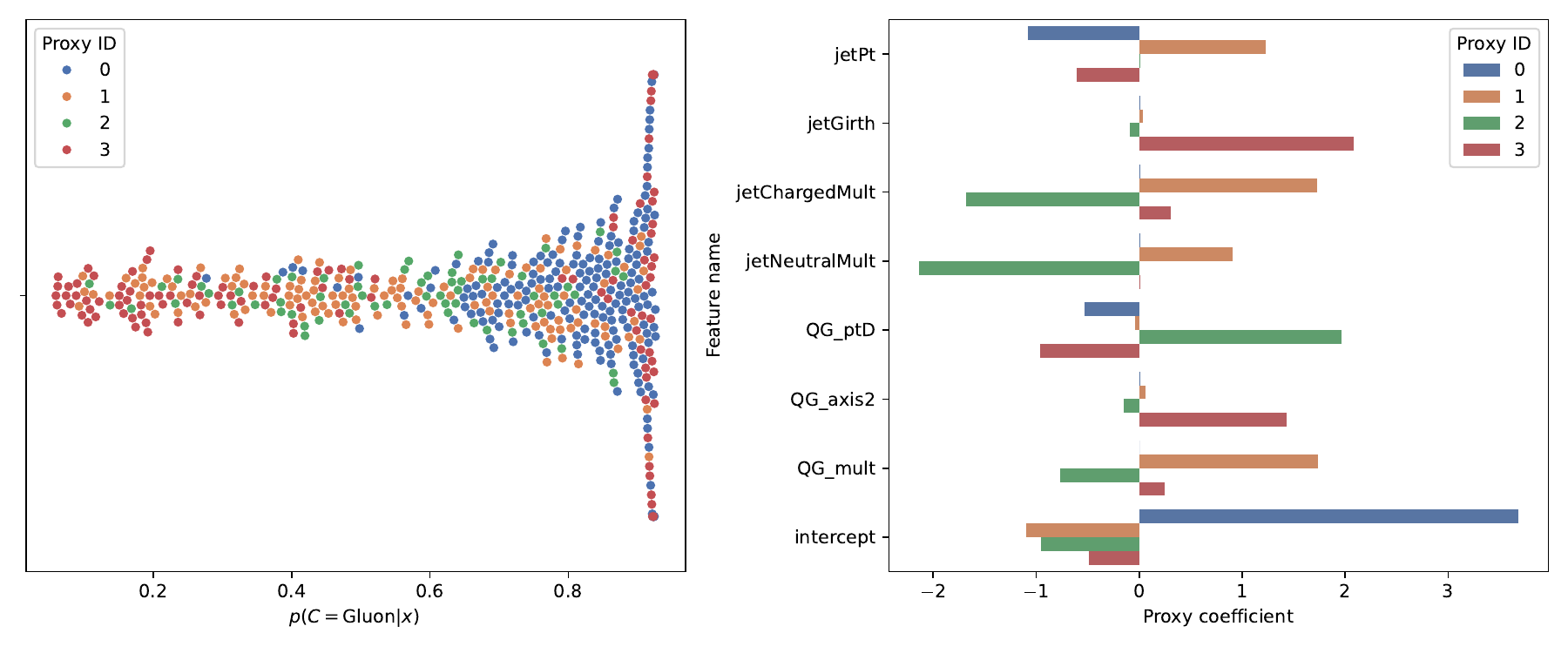}
    \caption{{\sc ExplainReduce} is able to find proxy sets that adhere to the underlying physical theory. The dataset consists of LHC proton-proton collision particle jets that are created by decaying gluons or quarks. We train a random forest classifier on the dataset and use {\sc slisemap} to generate 500 explanations, which are then reduced to $k=4$ proxies using the {\sc Balanced} reduction algorithm. The left panel shows a swarm plot of the 500 items sorted horizontally based on the predicted probability of corresponding to a gluon; the y-axis has no significance. The right panel shows the coefficients of the logistic regression proxy models, which match particle-physics-based intuition.}
    \label{fig:jets_example}
\end{figure}

We first trained a random forest model on the data and then applied {\sc slisemap} to find local explanations.
We clustered the {\sc slisemap} local explanations into 5 clusters, analysed the average models for each cluster, and found them to adhere well to physical theory.
When we generate 500 {\sc slisemap} local explanations on the same dataset and apply {\sc ExplainReduce} with the {\sc Balanced} reduction algorithm with $k=4$ proxies, we find the proxy set depicted in Fig. \ref{fig:jets_example}.
The left panel shows a swarm plot where each item is depicted on a horizontal line based on the probability predicted by the random forest for a jet corresponding to a gluon jet. It is coloured based on which proxy model is associated with the item.
The right panel shows the coefficients of the regularised logistic regression proxy models.
We find that the proxies are similar to the cluster mean models shown in the previous publication and show similar adherence to the underlying quantum chromodynamic theory \citep{cms2013performance}.
For example, wider jets (high {\sc jetGirth} and {\sc QG\_axis2}) are generally more gluon-like, and therefore these parameters are essential in classifying the jets.
Incidentally, proxy 3 (red) is associated with the most quark- and gluon-like jets and has high positive coefficients for both parameters.
Similarly, proxy 0 (blue) has negative coefficients for momentum ({\sc jetPt}) and {\sc QG\_ptD}, which measures the degree to which the total momentum parallels the jet.
Both of these features indicate a more quark-like jet.

\subsection{Comparison of reduction algorithms}
\subsubsection{Fidelity of proxy sets}
\begin{figure}
    \centering
    \includegraphics[width=\linewidth]{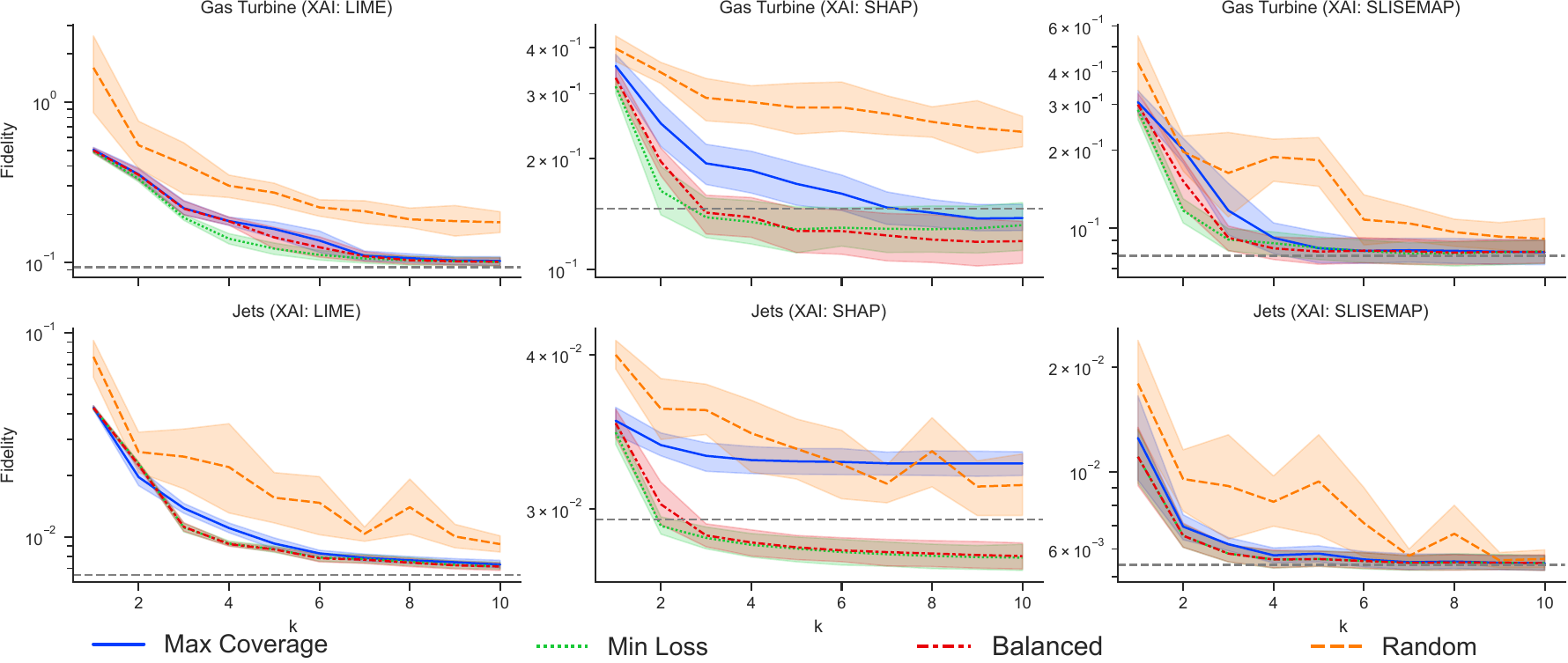}
    \caption{As few as $k=5$ proxies are able to reach nearly or better test set fidelity than the full set of local explanations. Lower values are better. The rows show different datasets, and the columns show different XAI generation methods. The different line colours and styles denote the reduction strategy, and the black horizontal line shows the performance of the full explanation set. The loss-minimising reduction methods, in particular, consistently reach a fidelity comparable to or even better than the full explanation set. Smaller fidelity is better.}
    \label{fig:fidelity_k}
\end{figure}

The interpretability of a set of local models as a global explanation can be directly linked to the size of that set.
As explained in a previous section, local explanation models often need to balance interpretability and granularity: a large set of local models may more easily cover a larger portion of the closed-box function, but the increased number of models makes it less understandable.
In Fig. \ref{fig:fidelity_k},\footnote{For full coverage on all of the twelve datasets, we refer to Appendix \ref{a:k_sensitivity_full}.} we show the test fidelity (as defined in Eq. \eqref{eq:fidelity}) as a function of the size of the reduced explanation set for a selection of datasets and local explanation algorithms, following the naming convention presented in Section \ref{sec:problem}.
The fidelity is calculated with respect to the closed-box predicted labels, with a fidelity of zero representing perfect adherence to the closed-box model.
As the figure shows, a set as small as $k=5$ proxies can reach or even surpass the fidelity of a set of $500$ local explanations for the selected datasets (horizontal black dashed line).
As expected, the two loss-minimising algorithms (green line for {\sc Min Loss} and red for {\sc Balanced}) have lower fidelity than the greedy coverage-maximising algorithm (blue line).
All three algorithms outperform the baseline of randomly picking explanations (yellow dotted line).
We also note that while the fidelity initially drops rapidly as $k$ increases, the metric quickly plateaus beyond $k>5$.
Therefore, it is enough for a user of {\sc ExplainReduce} to specify a reasonable value of $k$ for a faithful and interpretable global surrogate.

If a few proxies are enough to find a faithful representation, how many local explanations do we need to generate to find a good proxy set among them?
After all, the procedure may be infeasible if a very large set of initial local explanations is required.
For example, sampling-based XAI methods ({\sc lime}, {\sc shap}, and {\sc smoothgrad}) scale as $\mathcal{O}(nm)$, where $n = |\mathcal{D}|$ and $m$ is the number of samples per local model, which makes generating thousands of explanations expensive.

\begin{figure}
    \centering
    \includegraphics[width=\linewidth]{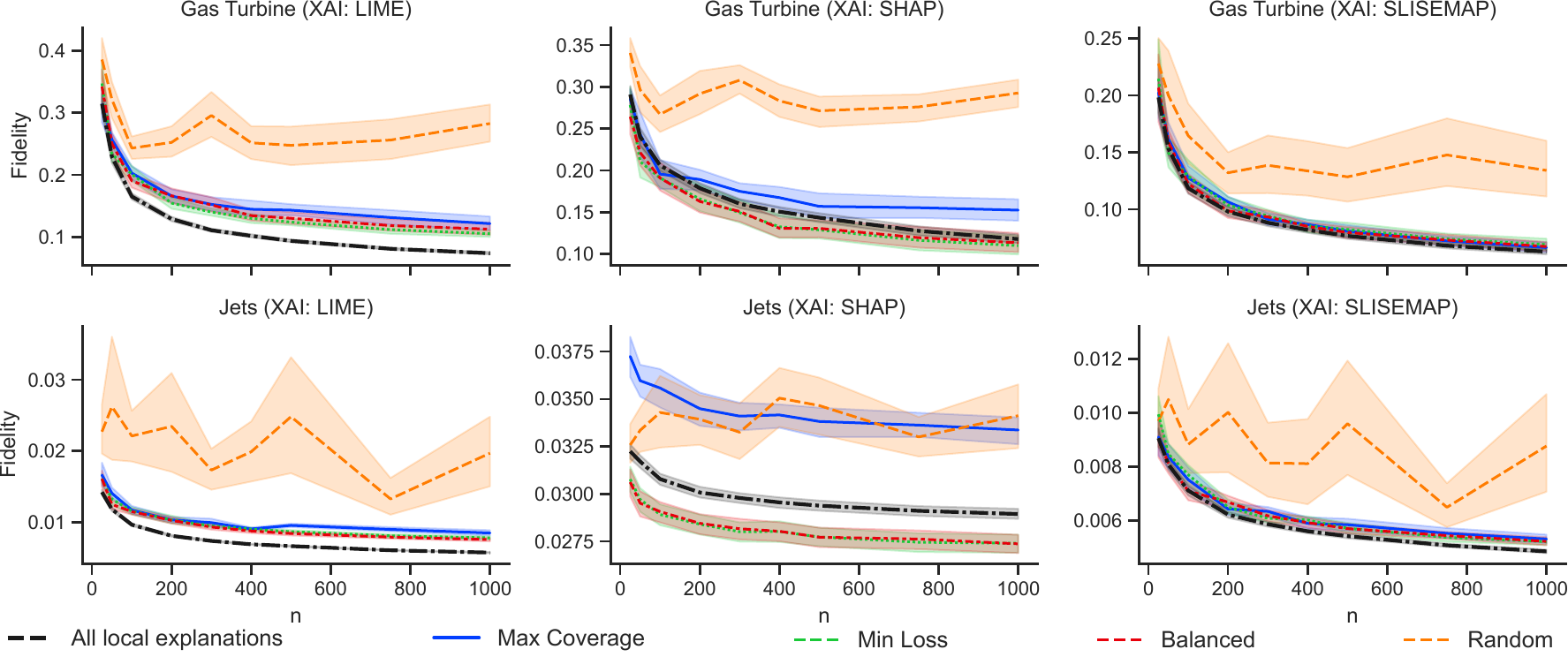}
    \caption{Test set fidelity quickly plateaus as the size of the initial local explanation set $n=|\mathcal{\bm{G}}|$ increases. Different datasets are depicted as rows, with the XAI methods used to produce the initial local explanations shown as rows. The coloured lines depict the test fidelity of the proxy sets produced with different reduction algorithms, with the thick black line denoting the performance of the entire local explanation set. Smaller fidelity is better.}
    \label{fig:fidelity_n}
\end{figure}

To study the effect of local model number to proxy set performance, we artificially limit the set of all local models $\bm{G}$ via random sampling between generation and reduction.
We then calculate the test fidelity of the proxy set trained on the limited universe, as described above, and compare that to the performance of the full explanation set.
As Fig. \ref{fig:fidelity_n} shows, the proxy sets generalise well even starting from a limited set of local models\footnote{Full version shown in Appendix \ref{a:fidelity_n_full}.}.
The coloured lines represent the various reduction algorithms, using the same colouring convention as in Fig. \ref{fig:fidelity_k}, while the thick dashed black line denotes the performance of the initial set of local explanations.
The loss-minimising greedy algorithms again achieve the best performance across the studied datasets and explanation methods, often matching and sometimes outperforming the full explanation set.
These results imply that {\sc ExplainReduce} does not require training local models for each item; instead, a randomly sampled subset of models suffices to find a good proxy set.
Based on these results, we set the default number of subsamples to $n=500$ in the remainder of this paper.

We also experimented with the impact of the hyperparameters $\varepsilon$, the loss threshold, and $\lambda$, the trade-off parameter. We found that the reduction algorithms maintain similar fidelity performance across a wide range of these hyperparameter values.
We refer to Appendix \ref{a:cov_eps_sensitivity} for detailed results.

\subsubsection{Coverage and stability}
\begin{figure}
    \centering
    \includegraphics[width=\linewidth]{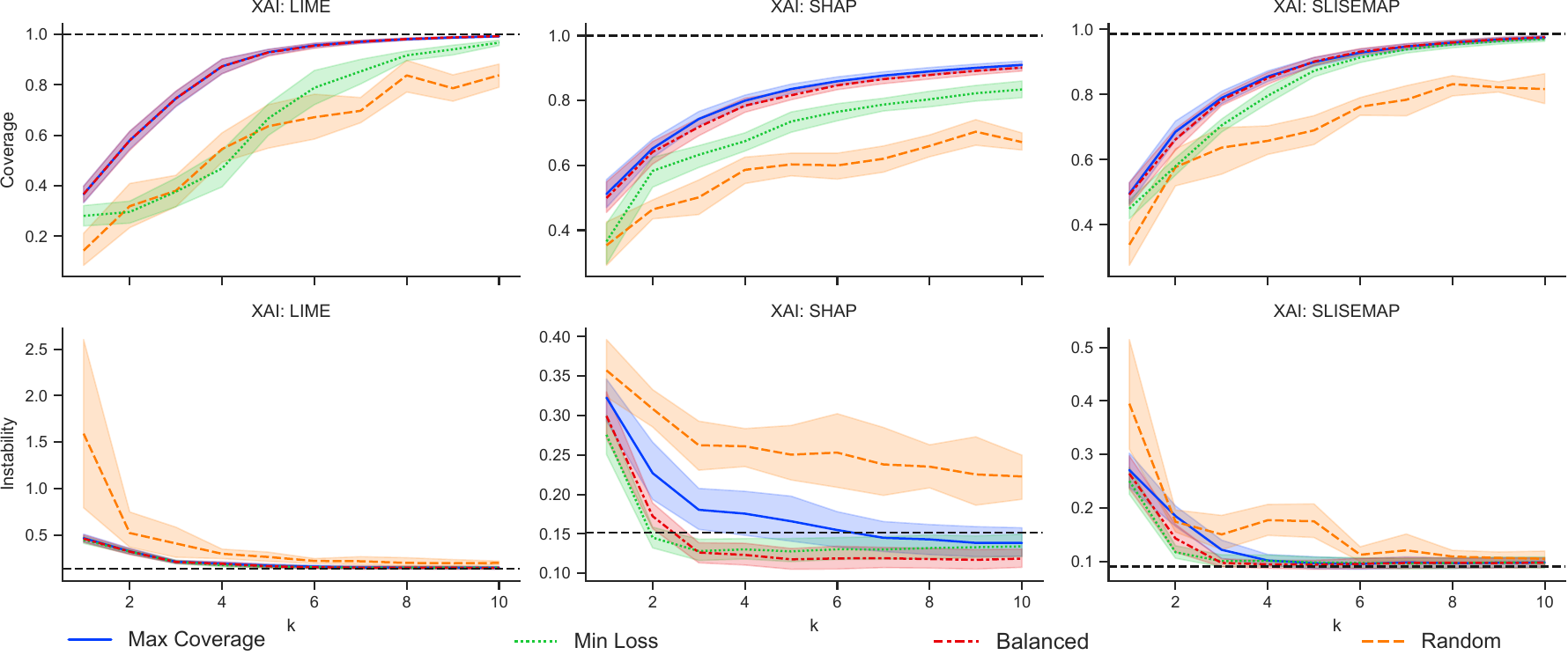}
    \caption{Small proxy sets also reach similar coverage and instability values as the full local explanation set. These results are from the Gas Turbine regression dataset. Naturally, the coverage-maximising algorithm shows best performance for coverage, but the balanced {\sc Balanced} algorithm has nearly equal results. Larger coverage and smaller instability are better.}
    \label{fig:coverage_stability}
\end{figure}

It should be no surprise that methods directly optimising for loss also show the best fidelity results.
However, an ideal global explanation method should be able to accurately explain most, if not all, of the items of interest.
Additionally, explanations are expected to be locally consistent, meaning that similar data items should generally have similar explanations. 
In Fig. \ref{fig:coverage_stability}, we show the evolution of training coverage and instability of reduction methods as a function of the proxy set size $k$ with respect to the closed-box predicted labels\footnote{Full versions shown in Appendix \ref{a:k_sensitivity_full}.}.
The results follow a similar pattern to the previous sections: a small number of proxies is enough to achieve high coverage and low instability, with optimisation-based approaches outperforming clustering-based alternatives.
Unsurprisingly, the coverage-optimising reduction method achieves the best coverage with the smallest number of proxies.
However, the loss-minimising algorithms perform surprisingly well, even compared to the coverage-optimising variant.
Notably, the balanced approach---the {\sc Balanced} reduction algorithm---almost matches the reduction algorithm directly optimising coverage.

The instability results show a similar trend to the fidelity results: the loss-minimising approaches have a slight edge over the coverage-maximising approach.

Combining the results from the fidelity, coverage, and instability experiments, the greedy {\sc Balanced} reduction algorithm appears to offer the best balance in performance across all three metrics.
The proxy set produced by this reduction method consistently achieves performance similar to, if not better than, the full explanation set, even with a few proxies.
Furthermore, these results and those presented in Appendix \ref{a:greedy} demonstrate more favourable practical approximation ratios than the worst-case performance discussed in Section \ref{sec:problem} for all three algorithms.

\subsection{Other global explanation and explanation aggregation methods}
\label{ssec:global_comp}

To verify the performance of the best {\sc ExplainReduce} algorithm, the {\sc Balanced} reduction, we compared the method to submodular pick of {\sc lime}, {\sc GLocalX} as well as the IP-based approach by \cite{li2022optimal} (here termed {\sc max spherical coverage} due to their definition of coverage through spherical neighbourhoods).
Note, however, that {\sc GLocalX} and the IP-based approach are limited to classification problems, and the submodular pick algorithm is limited to {\sc lime} only, unlike {\sc ExplainReduce}, which applies to the full gamut of tasks and XAI methods.
In Figure \ref{fig:global_fidelity_k}\footnote{Results for other datasets available in Appendix \ref{a:global_comparison}.}, we show the train and test fidelities and the train coverage for the model aggregation methods on the Adult dataset.
{\sc ExplainReduce} shows better or matching fidelity on both the train and test sets, showing that the method is able to find the most faithful subset of explanations both from the initial set of local models and also when generalising to novel items.
Similarly, our method achieves the highest or matching train set coverage, meaning that the subset found by our method has the highest likelihood on containing a reasonable local model for a given item.
It should also be noted that while {\sc ExplainReduce} and submodular pick are computationally very efficient and find the proxy set in a matter of seconds, the IP-based {\sc max spherical coverage} and {\sc GLocalX} algorithms take hundreds or thousands of seconds for the same task on the same hardware.
For some datasets, such as the Adult dataset depicted here, a single global linear model (horizontal red dashed line) has lower test set fidelity than even the full XAI local model sets with about $60\%$ coverage.
The local model sets and reductions outperform the single global model in train fidelity and coverage, meaning that more items in the test set can be associated with a local explanation within the set error tolerance.
As the rules produced by {\sc lore} are weaker predictors due to their simplicity, all the reduction methods are able to match the test set performance of the full rule set with very few rules.
Hence, we cannot say much about the relative performance of the methods beyond noting that {\sc ExplainReduce} generally performs comparatively to the rule-specific {\sc GLocalX}.

\begin{figure}
    \centering
    \includegraphics[width=\linewidth]{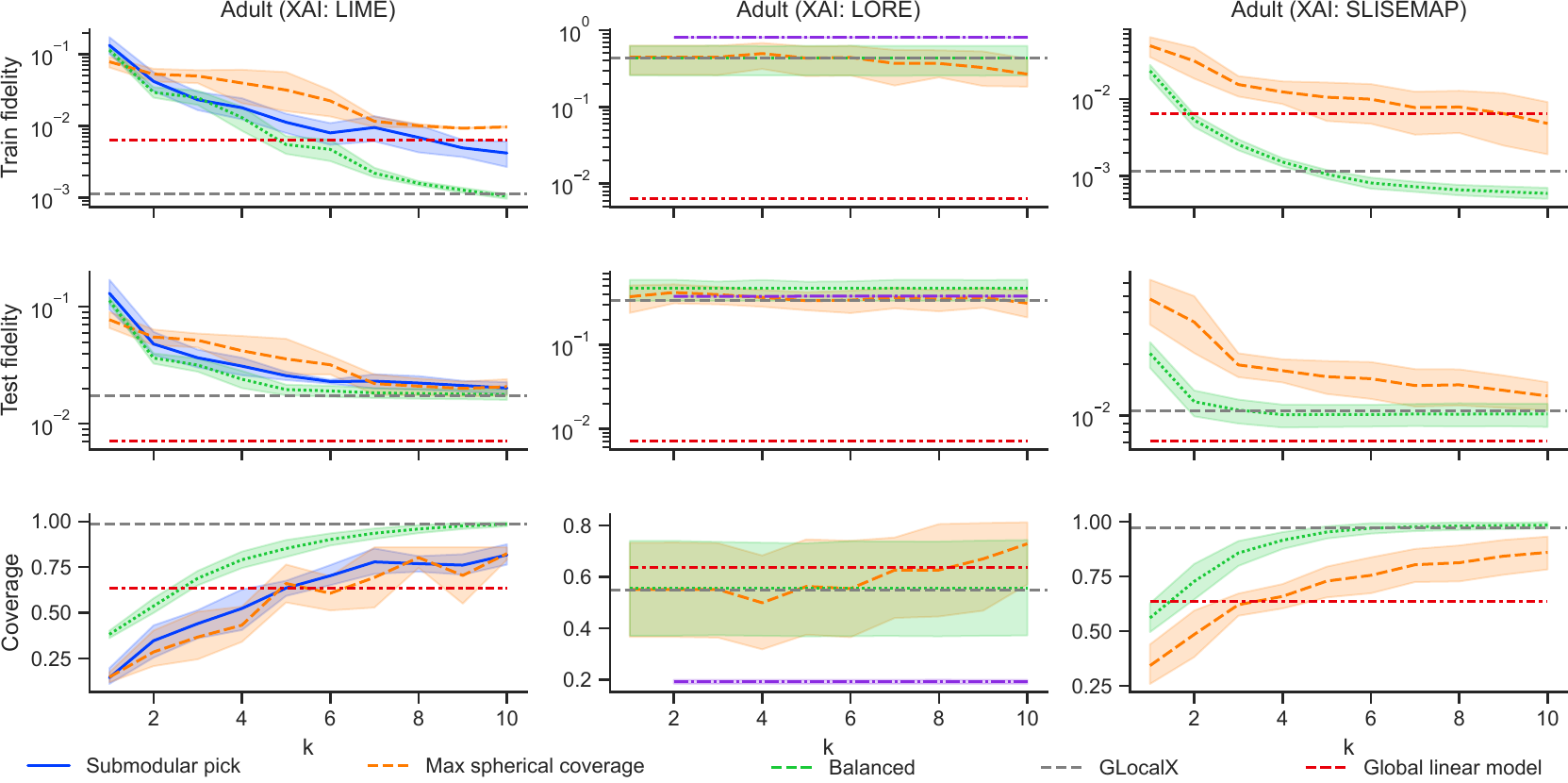}
    \caption{{\sc ExplainReduce}, using the {\sc Balanced} reduction algorithm (green dashed line), is able to match or exceed test set fidelity of other model aggregation methods. The black dashed line shows the performance of all local explanations. The red horizontal dashed line shows the fidelity of a single sparse global linear model, which, however, only covers about 60\% of the datapoints (bottom row). Smaller fidelity and higher coverage is better.}
    \label{fig:global_fidelity_k}
\end{figure}

\section{Discussion}

Using local surrogate models as explanations is a common approach for model-agnostic XAI methods.
In this paper, we have shown many local explanations are redundant, and that the {\sc ExplainReduce} procedure can find a post-hoc global explanation consisting of a small subset of simple models when provided a modest set of pre-trained local explanations.
The procedure is agnostic to both the underlying closed-box model and the XAI method, assuming that the XAI method can function as a generative (predictive) model.
These proxy sets achieve comparable fidelity, stability, and coverage to the full explanation set while remaining succinct enough to be easily interpretable by humans.
The {\sc Balanced} algorithm especially offers good performance in fidelity, coverage, and stability across different datasets.
Furthermore, we have shown how our approach shows favourable performance over other global explanation methods.
The procedure is not sensitive to hyperparameters, allowing for easy adoption.

The main limitation of our approach is that the presented implementation for the procedure simply finds a set of proxies that is (approximately) optimal from a global loss or coverage perspective.
In other words, there is no consideration of the spatial distribution of the data with respect to which proxy was associated with each item.
To deepen the insight provided by the method, an interesting idea would be to combine the reduction process with manifold learning techniques, enabling visualisation alongside the proxy set, similar to {\sc slisemap} and {\sc slipmap}.
For example, we might try to find an embedding where data items associated with a particular proxy would form continuous regions, significantly increasing interpretability and augmenting the method's capabilities as an exploratory data analysis tool.

Another interesting way to leverage the set of applicable local models is uncertainty quantification.
Instead of trying to find local models applicable to many data items, we could focus on all applicable models for a given item.
This set of applicable models could be considered as samples from the global distribution of explanations for that particular item.
Thus, we could use such sets (alongside the produced explanation) as a way to find, e.g., confidence intervals for explanations produced by a wide variety of XAI methods.

\bmhead{Acknowledgements}

We thank the Research Council of Finland for funding (decisions 364226, VILMA Centre for Excellence) and the Finnish Computing Competence Infrastructure (FCCI) for supporting this project with computational resources.

\bibliography{alternative2024}%

@misc{2019Gas,
  title = {Gas Turbine CO and NOx Emission Data Set},
  year = {2019},
  date = {2019},
  publisher = {UCI Machine Learning Repository},
  author= {UCI Machine Learning Repository},
  doi = {10.24432/C5WC95},
}

@article{agnar1994casebased,
  title = {Case-Based Reasoning: Foundational Issues, Methodological Variations, and System Approaches},
  author = {Aamodt, Agnar and Plaza, Enric},
  date = {1994-03},
  year = {1994},
  journaltitle = {Ai Communications},
  shortjournal = {AI Commun.},
  publisher = {IOS Press},
  location = {NLD},
  issn = {0921-7126},
  issue_date = {March 1994},
  pagetotal = {21},
  doi = {10.3233/aic-1994-7104}
}

@article{band2023medical,
  title = {Application of Explainable Artificial Intelligence in Medical Health: {{A}} Systematic Review of Interpretability Methods},
  author = {S Band, Shahab and Yarahmadi, Atefeh and Hsu, Chung-Chian and Biyari, Meghdad and Sookhak, Mehdi and Ameri, Rasoul and Dehzangi, Iman and Chronopoulos, Anthony Theodore and Liang, Huey-Wen},
  year = {2023},
  date = {2023},
  journaltitle = {Informatics in Medicine Unlocked},
  issn = {2352-9148},
  doi = {10.1016/j.imu.2023.101286},
}

@misc{birla2022vehicles,
  title = {Vehicle Dataset},
  author = {Birla, Nehal and Verma, Nishant and Kushawa, Nikhil},
  year = {2022},
  date = {2022},
  url = {https://www.kaggle.com/datasets/nehalbirla/vehicle-dataset-from-cardekho},
  version = {4}
}

@article{bjorklund2023slisemap,
  title = {{{SLISEMAP}}: {{Supervised}} Dimensionality Reduction through Local Explanations},
  shorttitle = {{{SLISEMAP}}},
  author = {Björklund, Anton and Mäkelä, Jarmo and Puolamäki, Kai},
  year = {2023},
  date = {2023-01},
  journaltitle = {Machine Learning},
  shortjournal = {Mach Learn},
  issn = {0885-6125, 1573-0565},
  doi = {10.1007/s10994-022-06261-1},
}

@inproceedings{bjorklund2024SLIPMAP,
  title = {{{SLIPMAP}}: {{Fast}} and~{{Robust Manifold Visualisation}} for~{{Explainable AI}}},
  shorttitle = {{{SLIPMAP}}},
  author = {Björklund, Anton and Seppäläinen, Lauri and Puolamäki, Kai},
  year = {2024},
  date = {2024},
  doi = {10.1007/978-3-031-58553-1_18},
  isbn = {978-3-031-58553-1}
}

@misc{CMS:opendata,
  title = {Simulated Data QCD\_Pt-15to3000\_TuneZ2star\_Flat\_8TeV\_pythia6 in {{AODSIM}} Format for 2012 Collision Data},
  author = {{CMS Collaboration}},
  year = {2017},
  date = {2017},
  publisher = {CERN Open Data Portal},
  url = {http://opendata.cern.ch/record/8882}
}

@misc{cms2013performance,
  title = {Performance of Quark/Gluon Discrimination in 8 {{TeV}} Pp Data},
  author = {{CMS collaboration}},
  year = {2013},
  date = {2013},
  number = {CMS-PAS-JME-13-002},
  institution = {CERN},
  location = {Geneva},
  url = {https://cds.cern.ch/record/1599732}
}

@inproceedings{dombrowski2019explanations,
  title = {Explanations Can Be Manipulated and Geometry Is to Blame},
  author = {Dombrowski, Ann-Kathrin and Alber, Maximillian and Anders, Christopher and Ackermann, Marcel and Müller, Klaus-Robert and Kessel, Pan},
  year = {2019},
  date = {2019},
  url = {\url{https://proceedings.neurips.cc/paper_files/paper/2019/file/bb836c01cdc9120a9c984c525e4b1a4a-Paper.pdf}}
}

@inproceedings{dosilovic2018survey,
  title = {Explainable Artificial Intelligence: {{A}} Survey},
  author = {Došilović, Filip Karlo and Brčić, Mario and Hlupić, Nikica},
  year = {2018},
  date = {2018},
  doi = {10.23919/MIPRO.2018.8400040}
}

@article{guidotti2018survey,
  title = {A Survey of Methods for Explaining Black Box Models},
  author = {Guidotti, Riccardo and Monreale, Anna and Ruggieri, Salvatore and Turini, Franco and Giannotti, Fosca and Pedreschi, Dino},
  year = {2018},
  date = {2018-08},
  journaltitle = {Acm Computing Surveys},
  shortjournal = {ACM Comput. Surv.},
  issn = {0360-0300},
  doi = {10.1145/3236009},
  articleno = {93},
  issue_date = {September 2019},
  pagetotal = {42}
}

@inproceedings{kim2016examples,
  title = {Examples Are Not Enough, Learn to Criticize! {{Criticism}} for {{Interpretability}}},
  author = {Kim, Been and Khanna, Rajiv and Koyejo, Oluwasanmi O},
  year = {2016},
  date = {2016},
  url = {\url{https://proceedings.neurips.cc/paper_files/paper/2016/file/5680522b8e2bb01943234bce7bf84534-Paper.pdf}}
}

@misc{laugel2018defining,
      title={Defining Locality for Surrogates in Post-hoc Interpretablity}, 
      author={Thibault Laugel and Xavier Renard and Marie-Jeanne Lesot and Christophe Marsala and Marcin Detyniecki},
      year={2018},
      eprint={1806.07498},
      archivePrefix={arXiv},
      primaryClass={cs.LG},
      url={https://arxiv.org/abs/1806.07498}, 
}

@article{li2022Optimal,
  title = {Optimal {{Local Explainer Aggregation}} for {{Interpretable Prediction}}},
  author = {Li, Qiaomei and Cummings, Rachel and Mintz, Yonatan},
  year = {2022},
  date = {2022-06-28},
  journaltitle = {Proceedings of the AAAI Conference on Artificial Intelligence},
  shortjournal = {AAAI},
  doi = {10.1609/aaai.v36i11.21458},
}

@inproceedings{lundberg2017unified,
  title = {A {{Unified Approach}} to {{Interpreting Model Predictions}}},
  author = {Lundberg, Scott M and Lee, Su-In},
  year = {2017},
  date = {2017},
  url = {\url{https://proceedings.neurips.cc/paper/2017/file/8a20a8621978632d76c43dfd28b67767-Paper.pdf}}
}

@incollection{mccormick2005submodular,
  title = {Submodular Function Minimization},
  author = {McCormick, S. Thomas},
  year = {2005},
  date = {2005},
  issn = {0927-0507},
  doi = {10.1016/S0927-0507(05)12007-6},
}

@article{nemhauser1978analysis,
  title = {An Analysis of Approximations for Maximizing Submodular Set Functions—{{I}}},
  author = {Nemhauser, G. L. and Wolsey, L. A. and Fisher, M. L.},
  year = {1978},
  date = {1978-12-01},
  journaltitle = {Mathematical Programming},
  shortjournal = {Mathematical Programming},
  issn = {1436-4646},
  doi = {10.1007/BF01588971},
}

@article{oikarinenDetectingVirtualConcept2021,
  title = {Detecting Virtual Concept Drift of Regressors without Ground Truth Values},
  author = {Oikarinen, Emilia and Tiittanen, Henri and Henelius, Andreas and Puolamäki, Kai},
  year = {2021},
  date = {2021-05},
  journaltitle = {Data Mining and Knowledge Discovery},
  shortjournal = {Data Min Knowl Disc},
  issn = {1384-5810, 1573-756X},
  doi = {10.1007/s10618-021-00739-7},
}

@article{peng2022industrial,
  title = {Industrial Big Data-Driven Mechanical Performance Prediction for Hot-Rolling Steel Using Lower Upper Bound Estimation Method},
  author = {Peng, Gongzhuang and Cheng, Yinliang and Zhang, Yufei and Shao, Jian and Wang, Hongwei and Shen, Weiming},
  year = {2022},
  date = {2022},
  journaltitle = {Journal of Manufacturing Systems},
  issn = {0278-6125},
  doi = {10.1016/j.jmsy.2022.08.014},
}

@misc{rajarshi2017life,
  title = {Life {{Expectancy}} ({{WHO}})},
  author = {Rajarshi, Kumar},
  date = {2017},
  year = {2017},
  url = {https://www.kaggle.com/datasets/kumarajarshi/life-expectancy-who},
  version = {1}
}

@article{ramakrishnan2014Quantum,
  title = {Quantum Chemistry Structures and Properties of 134 Kilo Molecules},
  author = {Ramakrishnan, Raghunathan and Dral, Pavlo O. and Rupp, Matthias and von Lilienfeld, O. Anatole},
  year = {2014},
  date = {2014-08-05},
  journaltitle = {Scientific Data},
  shortjournal = {Sci Data},
  issn = {2052-4463},
  doi = {10.1038/sdata.2014.22},
}

@inproceedings{ribeiro2016,
  title = {"{{Why Should I Trust You}}?": {{Explaining}} the {{Predictions}} of {{Any Classifier}}},
  shorttitle = {"{{Why Should I Trust You}}?},
  author = {Ribeiro, Marco Tulio and Singh, Sameer and Guestrin, Carlos},
  year = {2016},
  date = {2016-08-13},
  location = {San Francisco California USA},
  doi = {10.1145/2939672.2939778},
  eventtitle = {{{KDD}} '16: {{The}} 22nd {{ACM SIGKDD International Conference}} on {{Knowledge Discovery}} and {{Data Mining}}},
  isbn = {978-1-4503-4232-2}
}

@article{seppalainen2023using,
  doi = {10.1371/journal.pone.0297714},
  author = {Seppäläinen, Lauri AND Björklund, Anton AND Besel, Vitus AND Puolamäki, Kai},
  journal = {PLOS ONE},
  publisher = {Public Library of Science},
  title = {Using slisemap to interpret physical data},
  year = {2024} 
}

@article{setzu2021glocalx,
  title = {{{GLocalX}} - from Local to Global Explanations of Black Box {{AI}} Models},
  author = {Setzu, Mattia and Guidotti, Riccardo and Monreale, Anna and Turini, Franco and Pedreschi, Dino and Giannotti, Fosca},
  year = {2021},
  date = {2021},
  journaltitle = {Artificial Intelligence},
  issn = {0004-3702},
  doi = {10.1016/j.artint.2021.103457},
}

@inproceedings{shankar2019alime,
  title = {{{ALIME}}: {{Autoencoder}} Based Approach for Local Interpretability},
  author = {Shankaranarayana, Sharath M. and Runje, Davor},
  year = {2019},
  date = {2019},
  isbn = {978-3-030-33607-3},
  doi = {10.1007/978-3-030-33607-3_49}
}

@misc{slack2021reliable,
 author = {Slack, Dylan and Hilgard, Anna and Singh, Sameer and Lakkaraju, Himabindu},
 title = {Reliable Post hoc Explanations: Modeling Uncertainty in Explainability},
 url = {https://proceedings.neurips.cc/paper_files/paper/2021/file/4e246a381baf2ce038b3b0f82c7d6fb4-Paper.pdf},
 year = {2021}
}

@misc{smilkov2017smoothgrad,
      title={SmoothGrad: removing noise by adding noise}, 
      author={Daniel Smilkov and Nikhil Thorat and Been Kim and Fernanda Viégas and Martin Wattenberg},
      year={2017},
      eprint={1706.03825},
      archivePrefix={arXiv},
      primaryClass={cs.LG},
      url={https://arxiv.org/abs/1706.03825}, 
}

@misc{szegedy2014intriguing,
  title = {Intriguing Properties of Neural Networks},
  author = {Szegedy, Christian and Zaremba, Wojciech and Sutskever, Ilya and Bruna, Joan and Erhan, Dumitru and Goodfellow, Ian and Fergus, Rob},
  year = {2014},
  date = {2014},
  eprint = {1312.6199},
  eprinttype = {arXiv},
  eprintclass = {cs.CV},
  url = {https://arxiv.org/abs/1312.6199}
}

@misc{whiteson2014higgs,
  author       = {Whiteson, Daniel},
  title        = {{HIGGS}},
  year         = {2014},
  howpublished = {UCI Machine Learning Repository},
  doi         = {10.24432/C5V312}
}

@misc{zafar2019dlime,
  author    = {Muhammad Rehman Zafar and Naimul Mefraz Khan},
  title     = {DLIME: A Deterministic Local Interpretable Model-Agnostic Explanations Approach for Computer-Aided Diagnosis Systems},
  booktitle = {In proceeding of ACM SIGKDD Workshop on Explainable AI/ML (XAI) for Accountability, Fairness, and Transparency},
  year      = {2019},
  doi = {10.32920/22734359}
}

@inproceedings{zhao2021baylime,
  title = {{{BayLIME}}: {{Bayesian}} Local Interpretable Model-Agnostic Explanations},
  author = {Zhao, Xingyu and Huang, Wei and Huang, Xiaowei and Robu, Valentin and Flynn, David},
  year = {2021},
  date = {2021-07-27/2021-07-30},
  series = {Proceedings of Machine Learning Research},
  url = {\url{https://proceedings.mlr.press/v161/zhao21a.html}}
}

@article{guidotti2019lore,
  title = {Factual and Counterfactual Explanations for Black Box Decision Making},
  author = {Guidotti, Riccardo and Monreale, Anna and Giannotti, Fosca and Pedreschi, Dino and Ruggieri, Salvatore and Turini, Franco},
  date = {2019},
  year = {2019},
  journaltitle = {IEEE Intelligent Systems},
  doi = {10.1109/MIS.2019.2957223}
}

@incollection{yin2003cpar,
  title = {{{CPAR}}: {{Classification}} Based on Predictive Association Rules},
  author = {Yin, Xiaoxin and Han, Jiawei},
  eprint = {https://epubs.siam.org/doi/pdf/10.1137/1.9781611972733.40},
  doi = {10.1137/1.9781611972733.40},
  year={2003},
}

@misc{becker1996adult,
  title = {Adult dataset},
  author = {Becker, Barry and Kohavi, Ronny},
  date = {1996},
  year={1996},
  doi= {10.24432/C5XW20},
  organization = {UCI Repository of Machine Learning Database}
}

@misc{blake1998churn,
  title = {Churn {{Data Set}}},
  author = {Blake, C. L. and Merz, C. J.},
  date = {1998},
  year = {1998},
  url = {https://www.openml.org/search?type=data&sort=version&status=any&order=asc&exact_name=churn&id=40701},
  organization = {UCI Repository of Machine Learning Database}
}

@misc{hopkins1999spam,
  title = {Spambase},
  author = {Hopkins, Mark and Reeber, Erik and George, Forman and Suermondt, Jaap},
  date = {1999},
  year = {1999},
  organization = {UCI Repository of Machine Learning Database},
  doi = {10.24432/C53G6X},
}

@misc{bock2004telescope,
  title = {{{MAGIC}} Gamma Telescope},
  author = {Bock, R.},
  date = {2004},
  year = {2004},
  organization = {UCI Repository of Machine Learning Database},
  doi = {10.24432/C52C8B},
}

@article{dwivedi2023survey,
author = {Dwivedi, Rudresh and Dave, Devam and Naik, Het and Singhal, Smiti and Omer, Rana and Patel, Pankesh and Qian, Bin and Wen, Zhenyu and Shah, Tejal and Morgan, Graham and Ranjan, Rajiv},
title = {Explainable AI (XAI): Core Ideas, Techniques, and Solutions},
year = {2023},
issue_date = {September 2023},
publisher = {Association for Computing Machinery},
address = {New York, NY, USA},
issn = {0360-0300},
url = {https://doi.org/10.1145/3561048},
doi = {10.1145/3561048},
journal = {ACM Comput. Surv.},
month = jan,
articleno = {194},
numpages = {33}
}

@article{mersha2024survey,
  title = {Explainable Artificial Intelligence: {{A}} Survey of Needs, Techniques, Applications, and Future Direction},
  author = {Mersha, Melkamu and Lam, Khang and Wood, Joseph and AlShami, Ali K. and Kalita, Jugal},
  year = {2024},
  journal = {Neurocomputing},
  issn = {0925-2312},
  doi = {10.1016/j.neucom.2024.128111}
}

@article{rashid2024aisurvey,
  title = {{{AI}} Revolutionizing Industries Worldwide: {{A}} Comprehensive Overview of Its Diverse Applications},
  author = {Rashid, Adib Bin and Kausik, MD Ashfakul Karim},
  date = {2024},
  year = {2024},
  journaltitle = {Hybrid Advances},
  issn = {2773-207X},
  doi = {10.1016/j.hybadv.2024.100277},
  url = {https://www.sciencedirect.com/science/article/pii/S2773207X24001386}
}

@article{ali2023xai,
  title = {Explainable Artificial Intelligence ({{XAI}}): {{What}} We Know and What Is Left to Attain Trustworthy Artificial Intelligence},
  author = {Ali, Sajid and Abuhmed, Tamer and El-Sappagh, Shaker and Muhammad, Khan and Alonso-Moral, Jose M. and Confalonieri, Roberto and Guidotti, Riccardo and Del Ser, Javier and Díaz-Rodríguez, Natalia and Herrera, Francisco},
  date = {2023},
  year = {2023},
  journaltitle = {Information Fusion},
  issn = {1566-2535},
  doi = {10.1016/j.inffus.2023.101805},
  url = {https://www.sciencedirect.com/science/article/pii/S1566253523001148}
}

@book{molnar:2019:a,
  title = {Interpretable Machine Learning: A Guide for Making Black Box Models Interpretable},
  shorttitle = {Interpretable Machine Learning},
  author = {Molnar, Christoph},
  date = {2019},
  year = {2019},
  publisher = {Lulu},
  location = {Morrisville, North Carolina},
  address = {Morrisville, North Carolina},
  isbn = {978-0-244-76852-2},
  pagetotal = {314},
  doi = {10.1177/09726225241252009}
}

@article{burkhart2021survey,
  title = {A Survey on the Explainability of Supervised Machine Learning},
  author = {Burkart, Nadia and Huber, Marco F.},
  date = {2021-05},
  year = {2021},
  journaltitle = {J. Artif. Int. Res.},
  publisher = {AI Access Foundation},
  location = {El Segundo, CA, USA},
  issn = {1076-9757},
  doi = {10.1613/jair.1.12228},
  url = {https://doi.org/10.1613/jair.1.12228},
  issue_date = {May 2021},
  pagetotal = {73}
}

\newpage

\appendix

\section{Details on XAI methods used}
\label{a:xai}

In this section, we give an overview of the formulation of XAI methods analysed in this article.

Theoretically, the simplest way to produce a local surrogate model would be to calculate the gradient of the closed-box function.
This XAI method is often called {\sc vanillagrad} in the literature.
However, in practice, the gradients of machine learning models can be very noisy, as demonstrated by the effectiveness of adversarial attacks that exploit small perturbations in neural networks \citep{szegedy2014intriguing}.
On the other hand, many commonly used machine learning models, such as random forests, do not have well-defined gradients.
Hence, more involved approaches are warranted. 

{\sc smoothgrad} \citep{smilkov2017smoothgrad} attempts to solve the gradient noise problem in {\sc vanillagrad} by averaging over gradients sampled from the vicinity of the point under explanation.
Although the original paper only applies {\sc smoothgrad} to classification, the method can easily be extended to regression. 

Moving away from directly analysing the gradient, {\sc lime} \citep{ribeiro2016} and {\sc shap} \citep{lundberg2017unified} are perhaps the most widely known examples of practical local explanation generation methods.
In this paper, we focus on the {\sc kernel-shap} variant, which combines linear {\sc lime} with {\sc shap}.
Both {\sc lime} and {\sc kernel-shap} produce explanations in the form of additive linear models ($\Tilde{y} = \phi^T \bm{x}$).
Given an item $\bm{x}_i$, the methods sample novel items $\bm{x}'_i \in \mathcal{X}'$ in the neighbourhood of the first item, use the closed-box model to predict the labels for the novel items and find the linear model that best fits them \citep{lundberg2017unified},
\begin{equation}
    \hat{\phi} = \arg \min\nolimits_{\phi} \sum\nolimits_{\bm{x}'_i \in \mathcal{X}'} [f(\bm{x}'_i) - \phi^T \bm{x}'_i]^2 \pi_x(\bm{x}'_i) + \Omega(\phi),
\end{equation}
where $\pi_x$ represents a distance measure and $\Omega(\phi)$ is a regularisation term.
The difference between the methods lies in the choice of distance measure, which defines the notion of neighbourhood for $\bm{x}_i$; {\sc lime} most often uses either $L^2$ or cosine distance, whereas {\sc kernel-shap} utilises results in game theory \citep{slack2021reliable}.

{\sc smoothgrad}, {\sc lime} and {\sc shap} are based on sampling novel items, which, while simple to implement, introduce a unique set of challenges.
First, formulating a reliable data generation process or sampling scheme for all possible datasets is difficult, if not impossible \citep{guidotti2018survey, laugel2018defining}.
For example, images generated by adding random noise to the pixel values rarely resemble natural images.
Second, randomly generating new items might produce items that cannot occur naturally due to, for example, violating the laws of physics.
{\sc slisemap} \citep{bjorklund2023slisemap} and its variant, {\sc slipmap} \citep{bjorklund2024SLIPMAP}, produce both a low-dimensional embedding for visualisation and a local model for all training items without sampling any new points.
{\sc slisemap} finds both the embedding and local models by optimising a loss function consisting of two parts: an embedding term, where items with similar local explanations attract each other while repelling dissimilar ones and a local loss term for the explanation of each item:
\begin{equation}
    \min_{g_i} \mathcal{L}_i = \sum\nolimits_{i=1}^n \frac{\exp(-D(\bm{z}_i, \bm{z}_j))}{\sum_{k=1}^n \exp(-D(\bm{z}_k, \bm{z}_j))} \ell(g_i(\bm{x}_j), \bm{\hat{y}}_j) + \Omega(g_i)
\end{equation}
where $D(\cdot, \cdot)$ is the Euclidean distance in the embedding, $g_i$ represents the local model, $\ell$ is the local loss function, and $\Omega$ again denotes regularisation term(s).
In {\sc slisemap}, each item is fitted to its local models; in the {\sc slipmap} variant, the number of local models is fixed to some $p$, usually much less than the number of data items $n$, and the training items are mapped to one of the $p$ local models.

As an example of rule-based explanation method (to be paired with the {\sc GLocalX} aggregation method) we consider Local Rule-based Explanations ({\sc lore}) \citep{guidotti2019lore}.
Rule-based local explanations take the form of simple decision rules, such as $r = \{age\leq25; job=clerk; income\leq900\}\rightarrow deny$ for a loan application classification task.
In {\sc lore}, these rules are generated via sampling the neighbours of a given item using a genetic algorithm to explore the decision boundary around the item.
Then, the closed-box model is used as an oracle for the generated neighbourhood items data, and a decision tree is trained with the predicted labels, from which the resulting rule is derived.
The procedure is hence similar to generating {\sc lime} and {\sc kernel-shap} explanations.
The rules learned from the training data form a rule set.
To use a rule set to predict novel items, we followed the procedure used by the authors of the article proposing {\sc GLocalX} \citep{setzu2021glocalx} and outlined by \cite{yin2003cpar}.
We first find all rules applicable to a given novel item, and then use the rule with the highest accuracy on the test set among these to produce the prediction.

\newpage
\section{Dataset introduction}
\label{a:datasets}
Below, we describe each of the datasets used in the experiments in the paper.

\noindent\textit{Synthetic} \citep{bjorklund2023slisemap} is a synthetic regression dataset with $N$ data points, $M$ features, $k$ clusters, and cluster spread $s$.
Each cluster is associated with a linear regression model and a centroid.
Coefficient vectors $\beta_j \in \mathbb{R}^{M+1}$ and centroids $c_j \in \mathbb{R}^M$ are sampled from normal distributions, with repeated samplings if either the coefficients or the centroids are too similar between the clusters.
Each data point is assigned to a cluster $j_i$, with features $x_i$ sampled around $c_{j_i}$ and standardised.
The target variable is $y_i = x_i^\top \beta_{j_i} + \epsilon_i$, where $\epsilon_i$ is Gaussian noise with standard deviation $\sigma_e$.
In the experiment section, unless otherwise stated, we set the number of data points to $N = 5000$, the number of features to $M = 11$, the number of clusters to $k = 5$, and the standard deviation of Gaussian noise to $\sigma_e = 2.0$.
In the closed-box model, we use a random forest regressor.

\vspace{0.2cm}

\noindent\textit{Air Quality} \citep{oikarinenDetectingVirtualConcept2021} contains 7355 hourly observations of 12 different air quality measurements.
One of the measured qualities is chosen as the label, and the other values are used as covariates.
With this dataset, we use a random forest regressor as the closed-box model.

\vspace{0.2cm}

\noindent\textit{Life Expectancy} \citep{rajarshi2017life} comprises 2,938 instances of 22 distinct health-related measurements, spanning the years 2000 to 2015 across 193 countries.
The expected lifespan from birth in years is used as the dependent variable, while other features act as covariates.
The closed-box model we use with this dataset is a neural network.

\vspace{0.2cm}

\noindent\textit{Vehicle} \citep{birla2022vehicles} contains 2059 instances of 12 different car-related features, with the target being the resale value of the instance.
We use a support vector regressor as the closed-box model with the Vehicles dataset.

\vspace{0.2cm}

\noindent\textit{Gas Turbine} \citep{2019gas} is a regression dataset with 36,733 instances of 9 sensor measurements on a gas turbine to study gas emissions.
With this dataset, we used an Adaboost regression as the closed-box model.

\vspace{0.2cm}

\noindent\textit{QM9} \citep{ramakrishnan2014Quantum} is a regression dataset comprising 133,766 small organic molecules. Features are created with the Mordred molecular description calculator.%
The QM9 closed-box model is a neural network.

\vspace{0.2cm}

\noindent\textit{Adult} \citep{becker1996adult} is a classic classification dataset with 48,842 observations of individuals and 14 features based on census data. The goal is to predict whether or not an individual's annual income exceeds \$50,000.
The closed-box model we used was a neural network classifier.

\vspace{0.2cm}

\noindent\textit{Churn} \citep{blake1998churn} is a classification dataset with 5,000 instances and 20 features. The goal is to predict whether a customer of a telecommunications company left the services of that company (``churned'') using features that represent information about the customer and their telephone use.
With this dataset, we used a random forest closed-box model.

\vspace{0.2cm}

\noindent\textit{HIGGS} \citep{whiteson2014higgs} is a two-class classification dataset consisting of signal processes that produce Higgs bosons or are background.
The dataset contains nearly $100000$ instances with 28 features.
We used a gradient boosting classifier with this dataset as the closed-box model.

\vspace{0.2cm}

\noindent\textit{Jets} \citep{CMS:opendata} contains simulated LHC proton-proton collisions.
The collisions produce quarks and gluons that decay into cascades of stable particles called jets.
The classification task is to distinguish between jets generated by quarks and gluons. 
The dataset has $266421$ instances with 7 features.
The closed-box model we used with this dataset is a random forest classifier.

\vspace{0.2cm}

\noindent\textit{Spam} \citep{hopkins1999spam} or Spambase is a classification dataset with 4,601 instances and 57 features.
The dataset consists of various email metadata, and the labels correspond to whether or not the email is spam.
With this dataset, we used a gradient boosting classifier as the closed-box model.

\vspace{0.2cm}

\noindent\textit{Telescope} \citep{bock2004telescope} is a classification dataset with 13,376 instances with 10 features.
The data are Monte Carlo simulated instances of Cherenkov radiation produced by high-energy gamma rays hitting the atmosphere, as they would be detected by a ground-based atmospheric Cherenkov gamma telescope.
The goal is to classify high-energy gamma events from background observations.
The closed-box model we used with this dataset was a neural network classifier.

\clearpage  %

\section{Fidelity, coverage and stability of the proxy sets as a function of proxy number}\label{a:k_sensitivity_full}
\begin{figure}
    \centering
    \includegraphics[width=\linewidth]{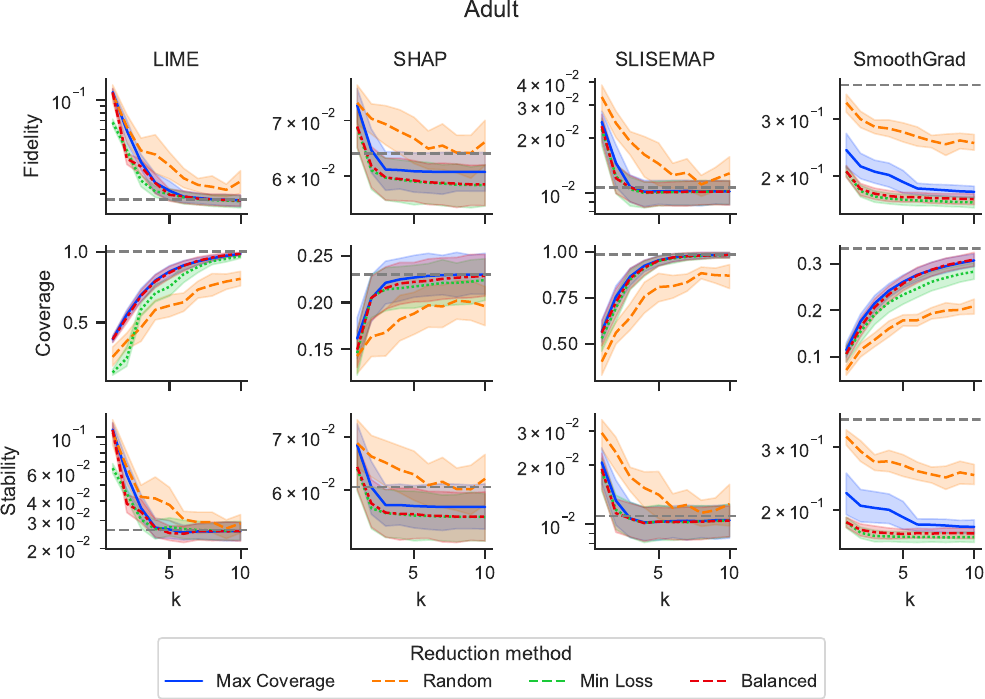}
    \caption{Adult dataset reduction function comparison as a function of $k$ on fidelity, coverage and stability.}
    \label{fig:ks_adult}
\end{figure}

\begin{figure}
    \centering
    \includegraphics[width=\linewidth]{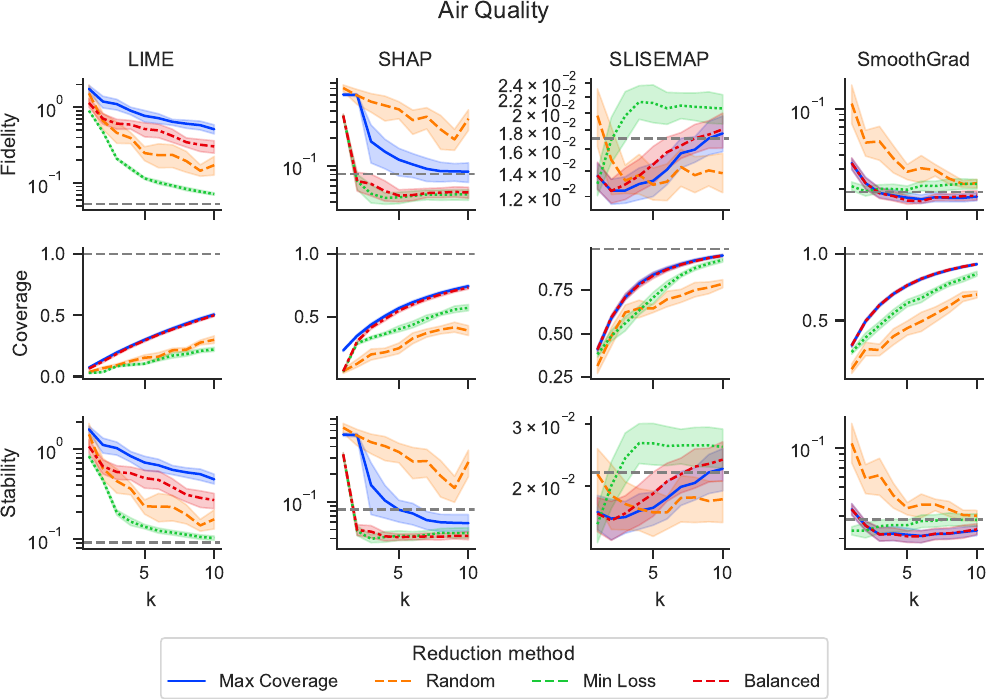}
    \caption{Air Quality dataset reduction function comparison as a function of $k$ on fidelity, coverage and stability.}
    \label{fig:ks_aq}
\end{figure}

\begin{figure}
    \centering
    \includegraphics[width=\linewidth]{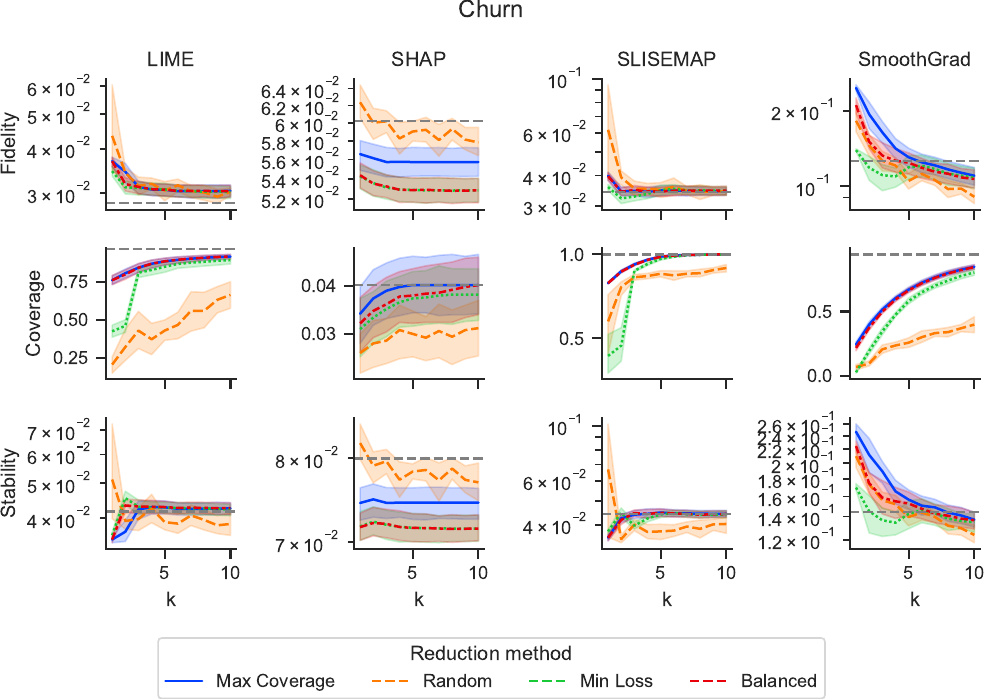}
    \caption{Churn dataset reduction function comparison as a function of $k$ on fidelity, coverage and stability.}
    \label{fig:ks_churn}
\end{figure}

\begin{figure}
    \centering
    \includegraphics[width=\linewidth]{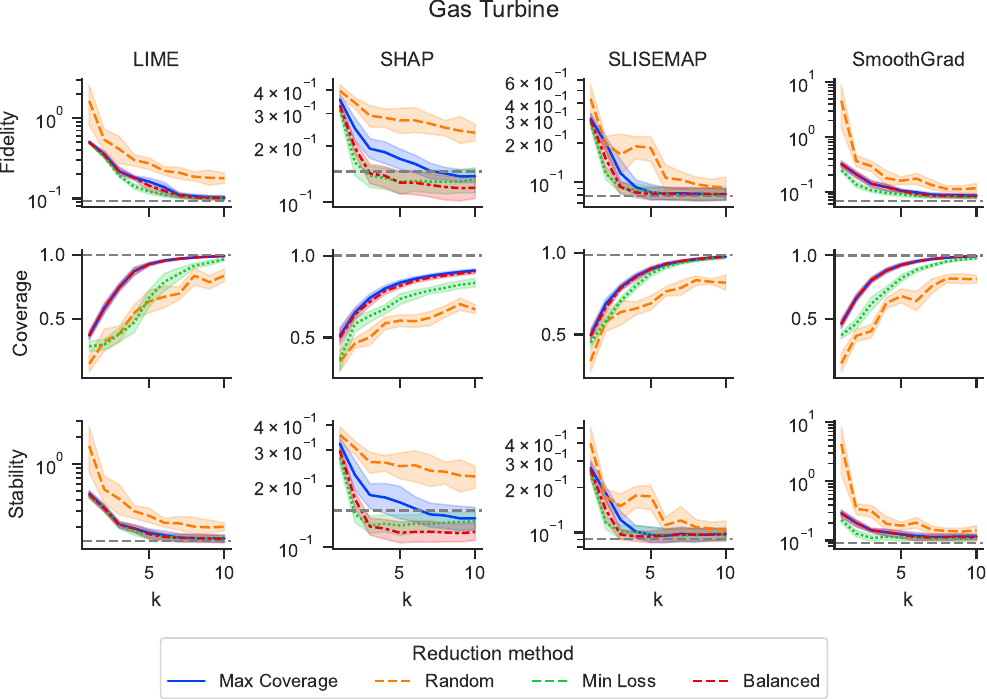}
    \caption{Gas Turbine dataset reduction function comparison as a function of $k$ on fidelity, coverage and stability.}
    \label{fig:ks_gt}
\end{figure}

\begin{figure}
    \centering
    \includegraphics[width=\linewidth]{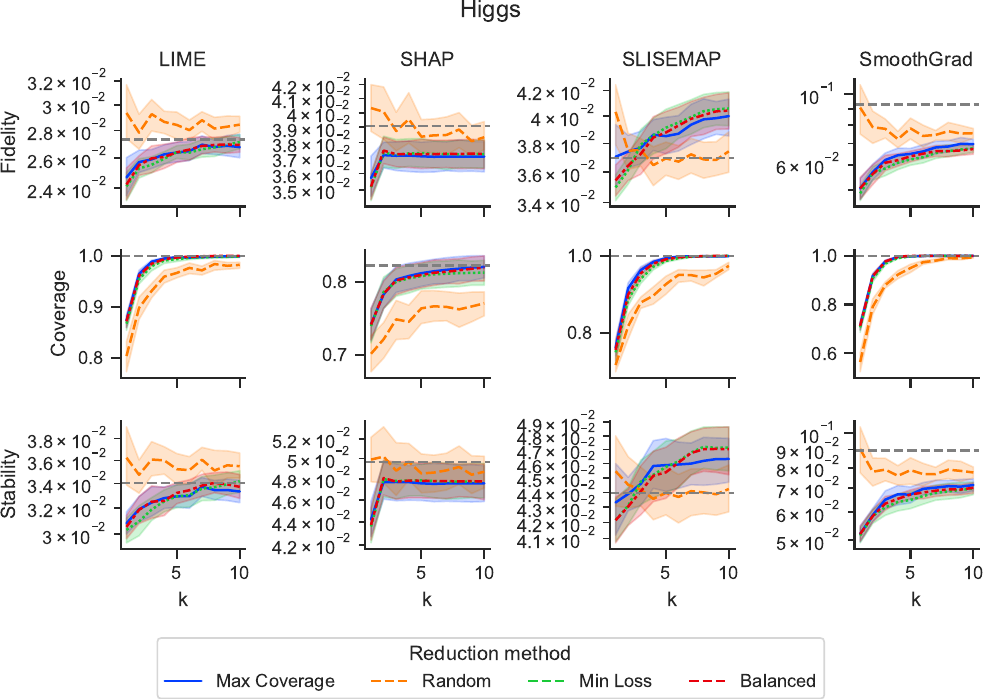}
    \caption{Higgs dataset reduction function comparison as a function of $k$ on fidelity, coverage and stability.}
    \label{fig:ks_higgs}
\end{figure}

\begin{figure}
    \centering
    \includegraphics[width=\linewidth]{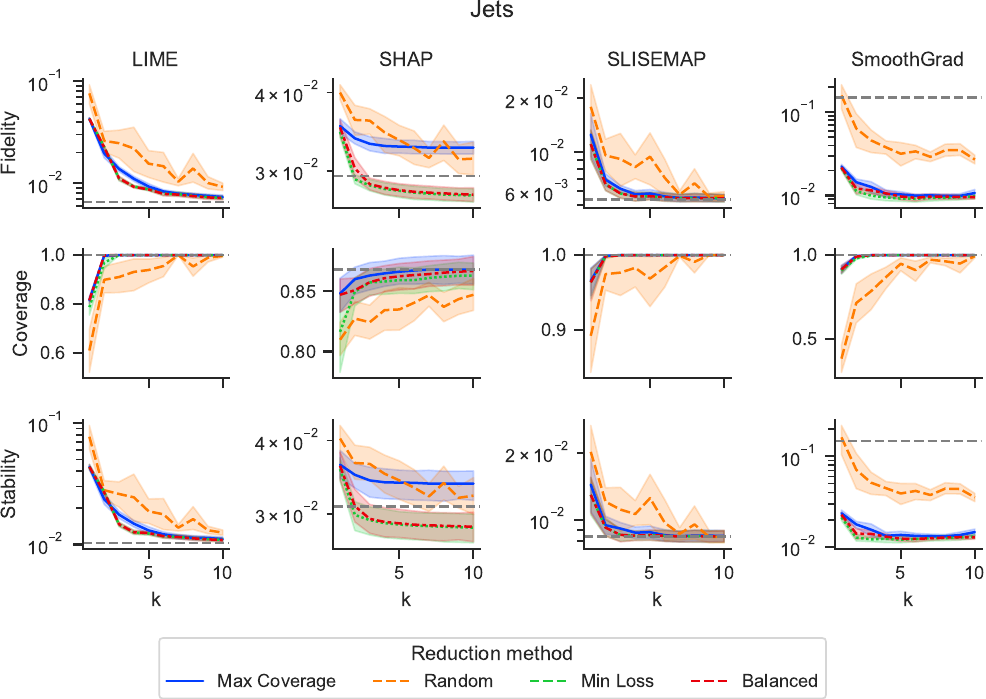}
    \caption{Jets dataset reduction function comparison as a function of $k$ on fidelity, coverage and stability.}
    \label{fig:ks_jets}
\end{figure}

\begin{figure}
    \centering
    \includegraphics[width=\linewidth]{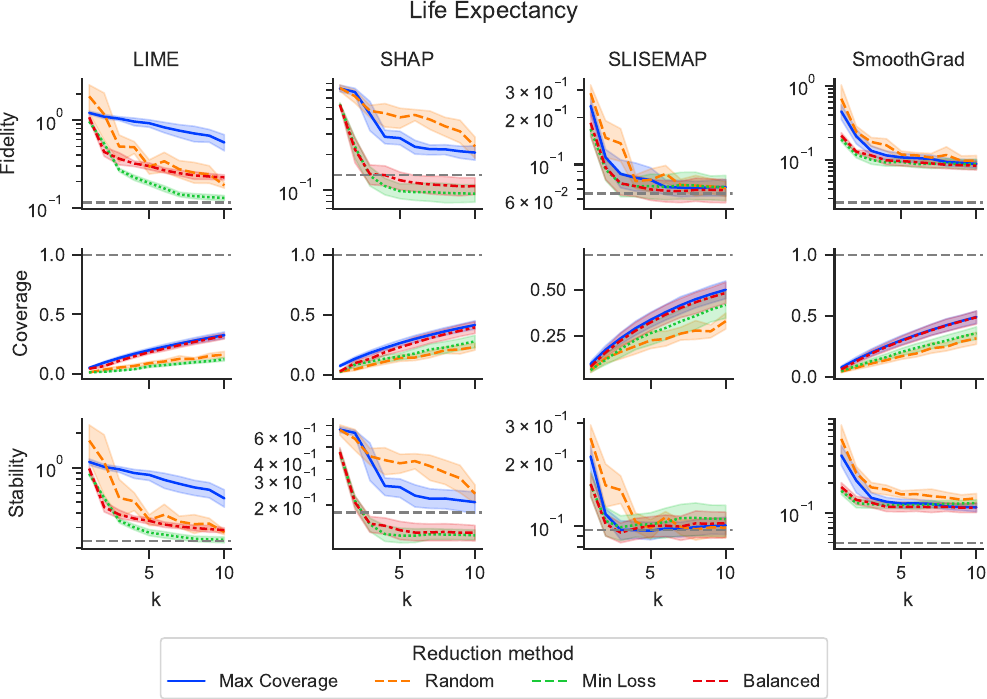}
    \caption{Life Expectancy dataset reduction function comparison as a function of $k$ on fidelity, coverage and stability.}
    \label{fig:ks_le}
\end{figure}

\begin{figure}
    \centering
    \includegraphics[width=\linewidth]{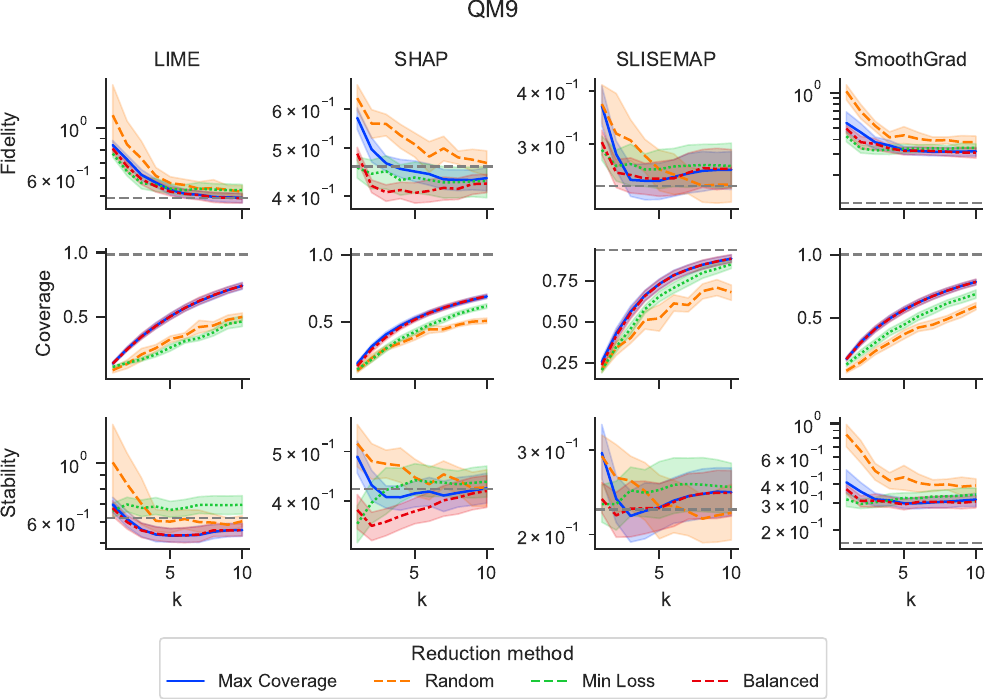}
    \caption{QM9 dataset reduction function comparison as a function of $k$ on fidelity, coverage and stability.}
    \label{fig:ks_qm9}
\end{figure}

\begin{figure}
    \centering
    \includegraphics[width=\linewidth]{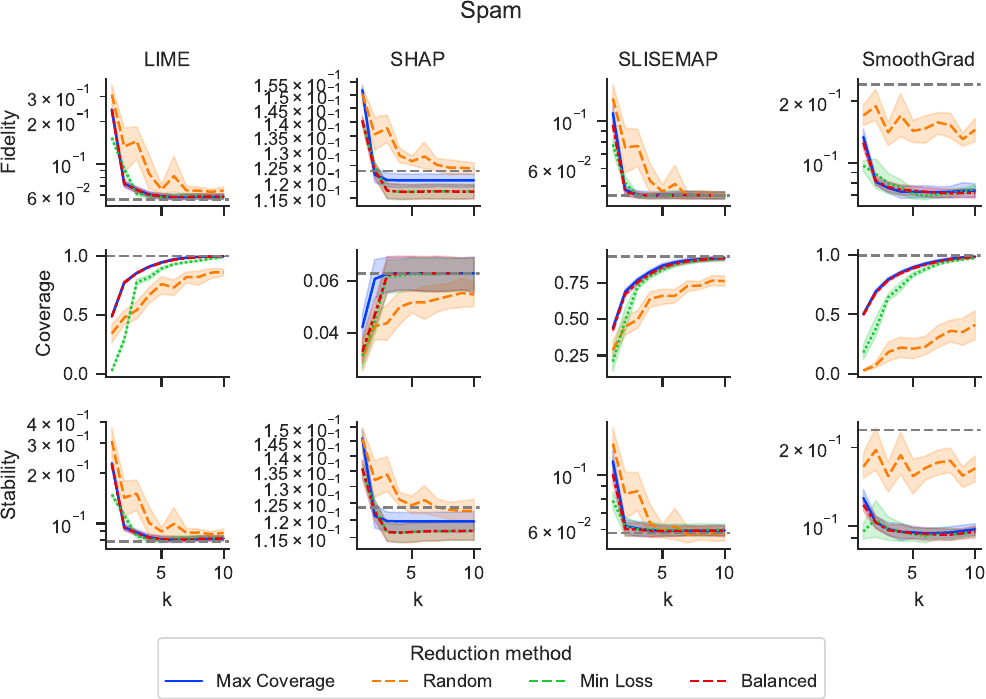}
    \caption{Spam dataset reduction function comparison as a function of $k$ on fidelity, coverage and stability.}
    \label{fig:ks_spam}
\end{figure}

\begin{figure}
    \centering
    \includegraphics[width=\linewidth]{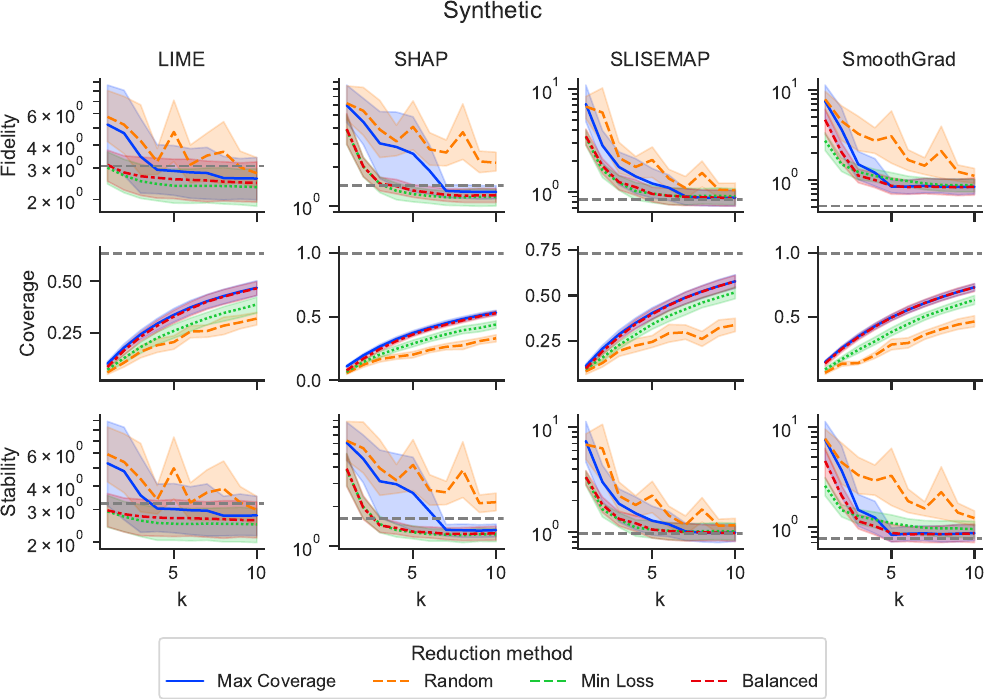}
    \caption{Synthetic dataset reduction function comparison as a function of $k$ on fidelity, coverage and stability.}
    \label{fig:ks_synth}
\end{figure}

\begin{figure}
    \centering
    \includegraphics[width=\linewidth]{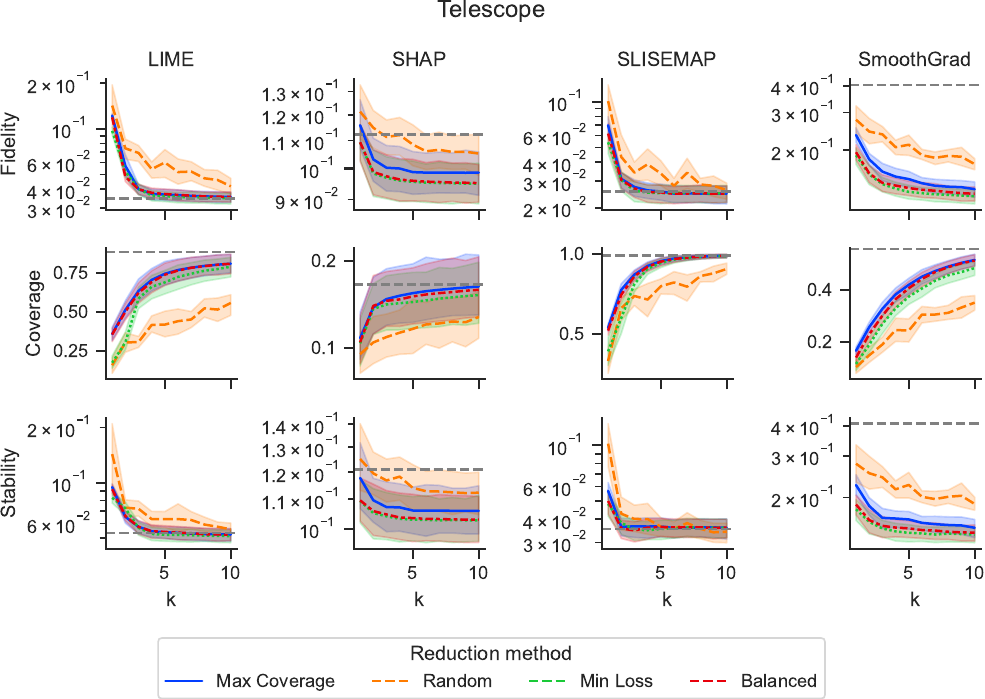}
    \caption{Telescope dataset reduction function comparison as a function of $k$ on fidelity, coverage and stability.}
    \label{fig:ks_telescope}
\end{figure}

\begin{figure}
    \centering
    \includegraphics[width=\linewidth]{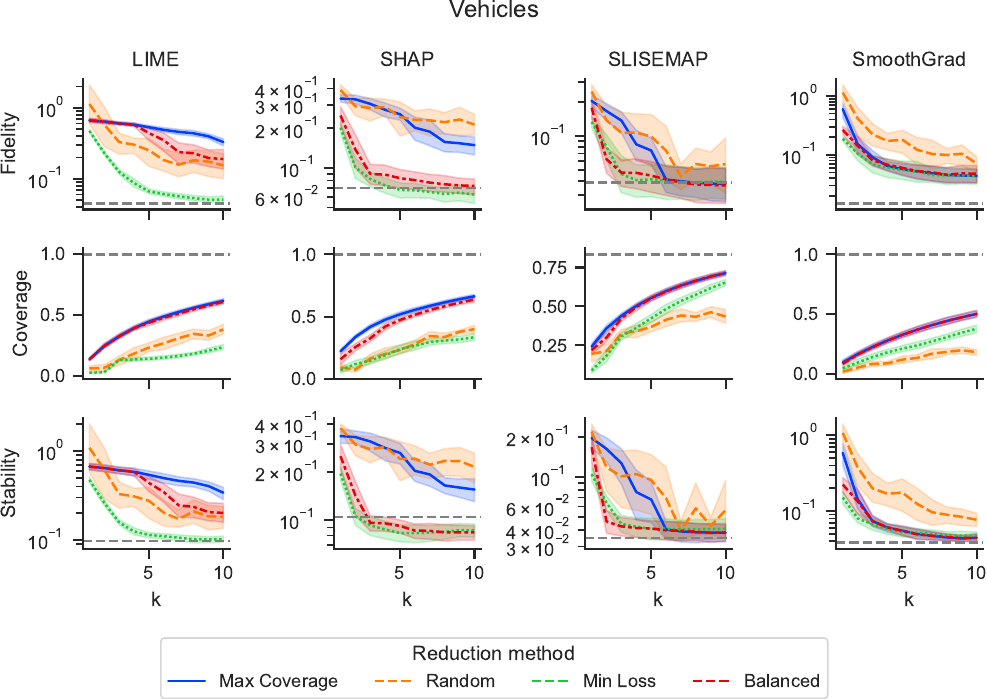}
    \caption{Vehicles dataset reduction function comparison as a function of $k$ on fidelity, coverage and stability.}
    \label{fig:ks_vehicles}
\end{figure}

Figures \ref{fig:ks_adult}-\ref{fig:ks_vehicles} show the complete fidelity, coverage and stability plots as a function of $k$. Each figure corresponds to a different dataset, and each column represents a local model generation method.
{\sc ExplainReduce} achieves strong generalisation across all datasets.
The loss-minimisation algorithms outperform or match coverage-optimising approaches, while remaining computationally efficient.
Although {\sc slisemap} often provides the most stable local explanations, the results indicate that the reduction algorithm—rather than the explainer alone—determines the quality of the global surrogate.
The combination of {\sc slisemap} with the Greedy Minimum loss reduction method consistently yields low test-set loss and high coverage.
In most datasets, as few as five proxy models ($k \approx 5$) suffice to approximate or even surpass the fidelity of the full ensemble of local explanations.
These results confirm that ExplainReduce can drastically compress hundreds of local models into a handful of interpretable surrogates while retaining predictive consistency.

\FloatBarrier  %
\clearpage  %

\section{Fidelity of the proxy sets as a function of subsample size}\label{a:fidelity_n_full}
\begin{figure}
    \centering
    \includegraphics[width=\linewidth]{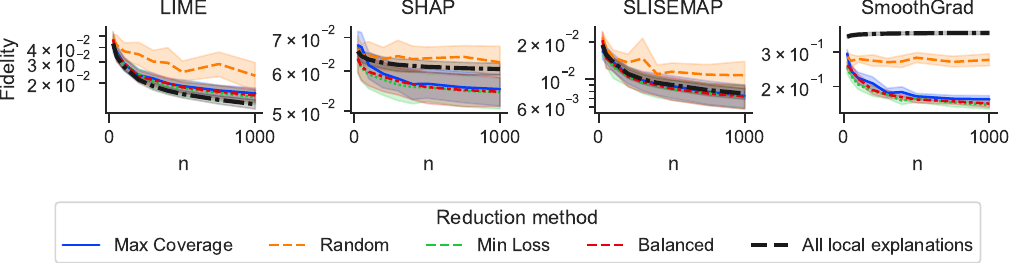}
    \caption{Adult dataset reduction function comparison as a function of $n$ on fidelity.}
    \label{fig:ns_adult}
\end{figure}

\begin{figure}
    \centering
    \includegraphics[width=\linewidth]{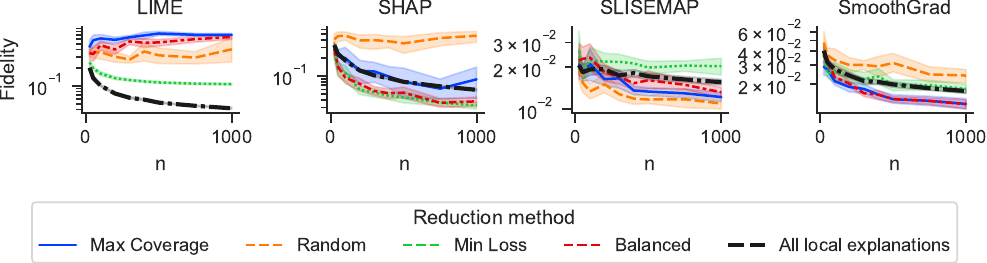}
    \caption{Air Quality dataset reduction function comparison as a function of $n$ on fidelity.}
    \label{fig:ns_aq}
\end{figure}

\begin{figure}
    \centering
    \includegraphics[width=\linewidth]{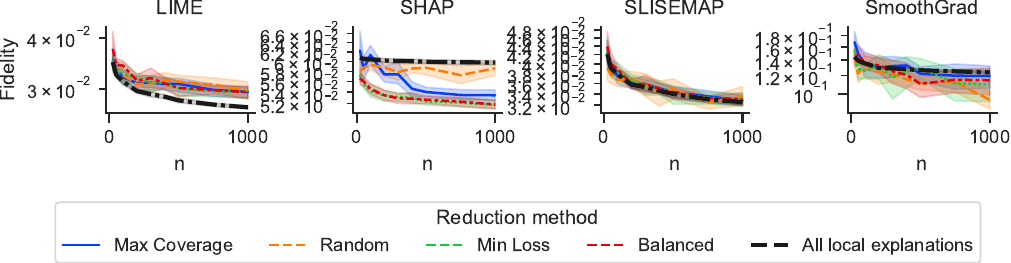}
    \caption{Churn dataset reduction function comparison as a function of $n$ on fidelity.}
    \label{fig:ns_churn}
\end{figure}

\begin{figure}
    \centering
    \includegraphics[width=\linewidth]{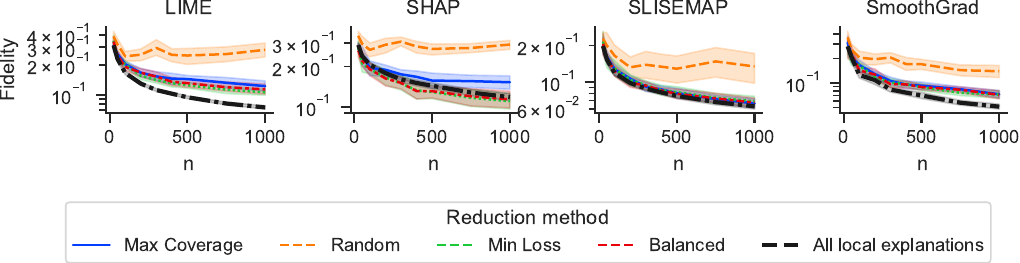}
    \caption{Gas Turbine dataset reduction function comparison as a function of $n$ on fidelity.}
    \label{fig:ns_gt}
\end{figure}

\begin{figure}
    \centering
    \includegraphics[width=\linewidth]{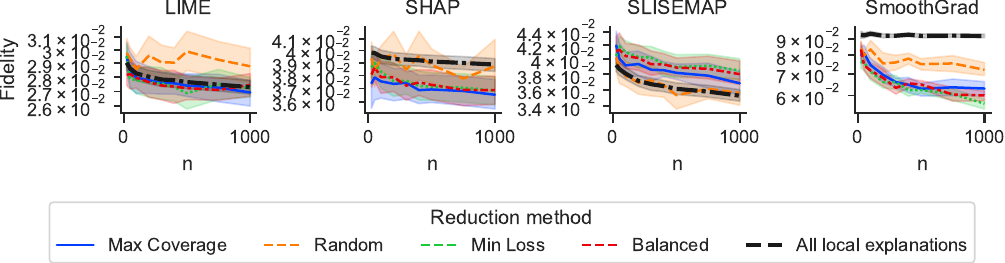}
    \caption{Higgs dataset reduction function comparison as a function of $n$ on fidelity.}
    \label{fig:ns_higgs}
\end{figure}

\begin{figure}
    \centering
    \includegraphics[width=\linewidth]{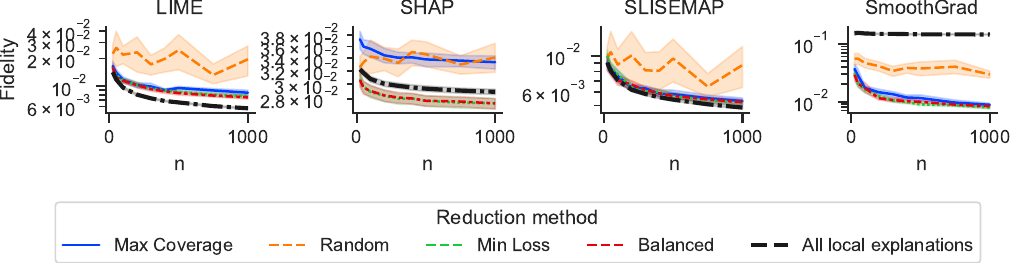}
    \caption{Jets dataset reduction function comparison as a function of $n$ on fidelity.}
    \label{fig:ns_jets}
\end{figure}

\begin{figure}
    \centering
    \includegraphics[width=\linewidth]{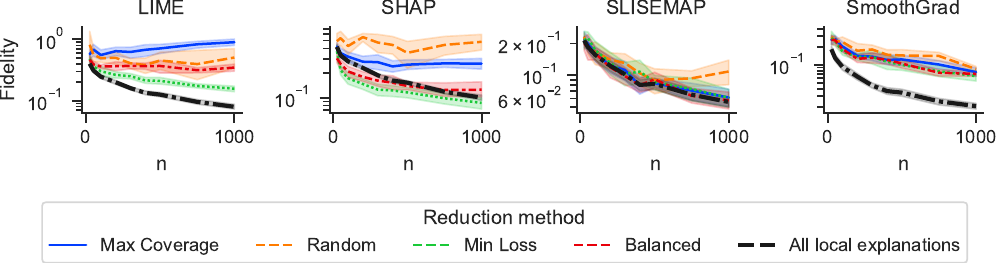}
    \caption{Life Expectancy dataset reduction function comparison as a function of $n$ on fidelity.}
    \label{fig:ns_le}
\end{figure}

\begin{figure}
    \centering
    \includegraphics[width=\linewidth]{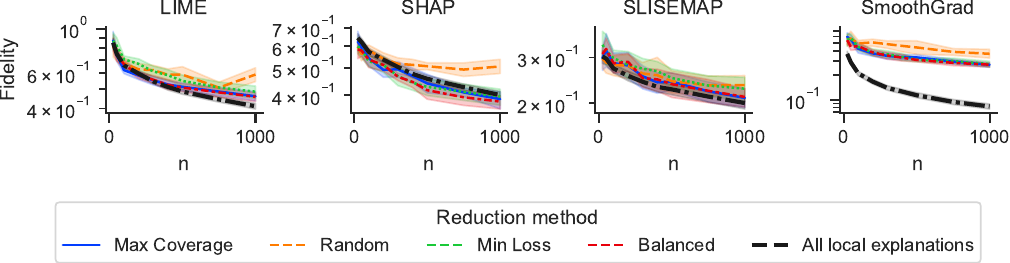}
    \caption{QM9 dataset reduction function comparison as a function of $n$ on fidelity.}
    \label{fig:ns_qm9}
\end{figure}

\begin{figure}
    \centering
    \includegraphics[width=\linewidth]{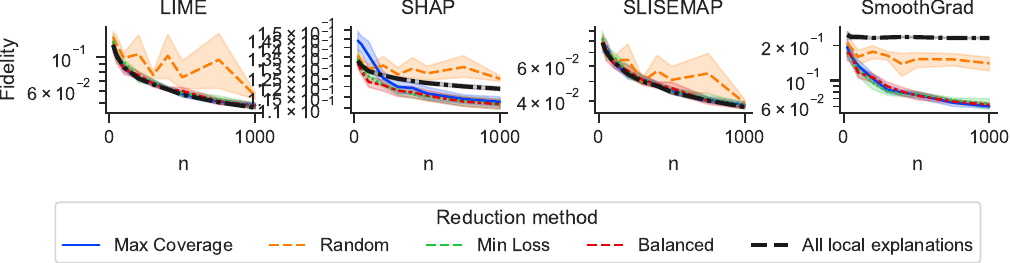}
    \caption{Spam dataset reduction function comparison as a function of $n$ on fidelity.}
    \label{fig:ns_spam}
\end{figure}

\begin{figure}
    \centering
    \includegraphics[width=\linewidth]{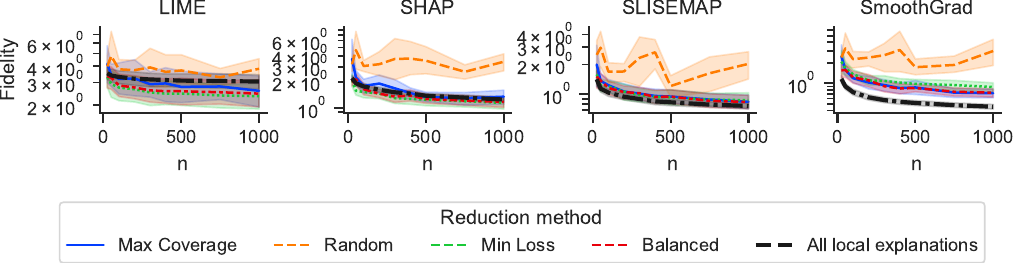}
    \caption{Sythetic dataset reduction function comparison as a function of $n$ on fidelity.}
    \label{fig:ns_synth}
\end{figure}

\begin{figure}
    \centering
    \includegraphics[width=\linewidth]{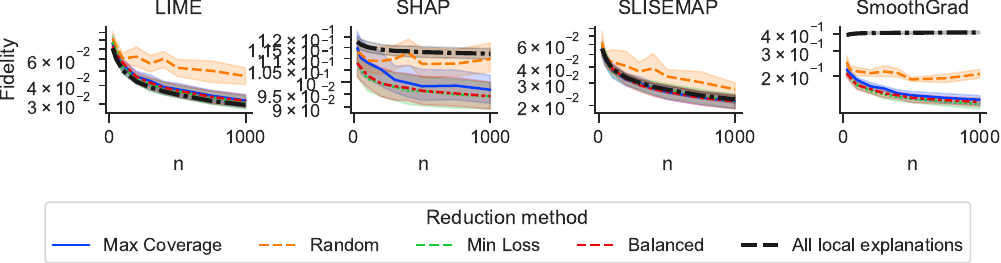}
    \caption{Telescope dataset reduction function comparison as a function of $n$ on fidelity.}
    \label{fig:ns_telescope}
\end{figure}

\begin{figure}
    \centering
    \includegraphics[width=\linewidth]{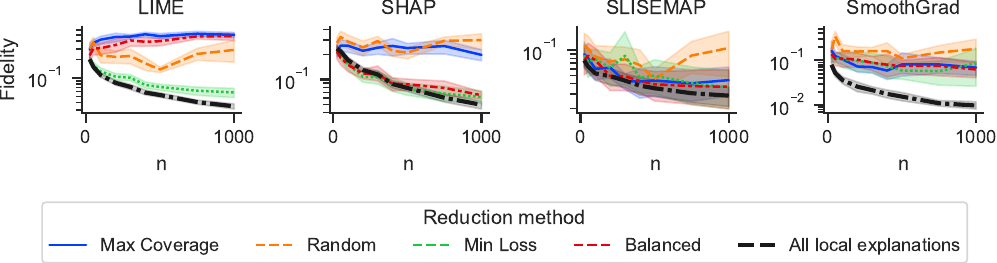}
    \caption{Vehicles dataset reduction function comparison as a function of $n$ on fidelity.}
    \label{fig:ns_vehicles}
\end{figure}

Figures \ref{fig:ns_adult}-\ref{fig:ns_vehicles} show the complete fidelity plots as a function of subsample size at proxy set size $k=5$. Each Figure corresponds to a different dataset, and each column represents a local model generation method. A general trend emerges, indicating improved fidelity with increasing subsample size.

\FloatBarrier  %
\clearpage  %

\section{Sensitivity analysis}\label{a:cov_eps_sensitivity}
 
Many of the reduction methods outlined in this paper require the user to select an error threshold $\varepsilon$, with the {\sc Balanced} algorithm requiring an additional trade-off parameter $\lambda$.
Since $\varepsilon$ is a parameter in computing coverage, its value can significantly impact the performance of the {\sc ExplainReduce} procedure.
In this section, we study the sensitivity of the procedure to these hyperparameters.

As the error tolerance $\varepsilon$ is defined in comparison to the absolute loss, the exact value of the parameter is task- and data-dependent.
In this section, we set the error tolerance as a quantile of the training loss of the underlying closed-box function $f$, i.e., $\varepsilon = q(\bm{L}_{BB}, p)$, where $q$ denotes the percentile function, $\bm{L}_{BB, i} = \ell(f(\bm{x}_i), \bm{\hat{y}}_i)$ and $p$ is the percentile value.

\begin{figure}[ht]
    \centering
    \includegraphics[width=\linewidth]{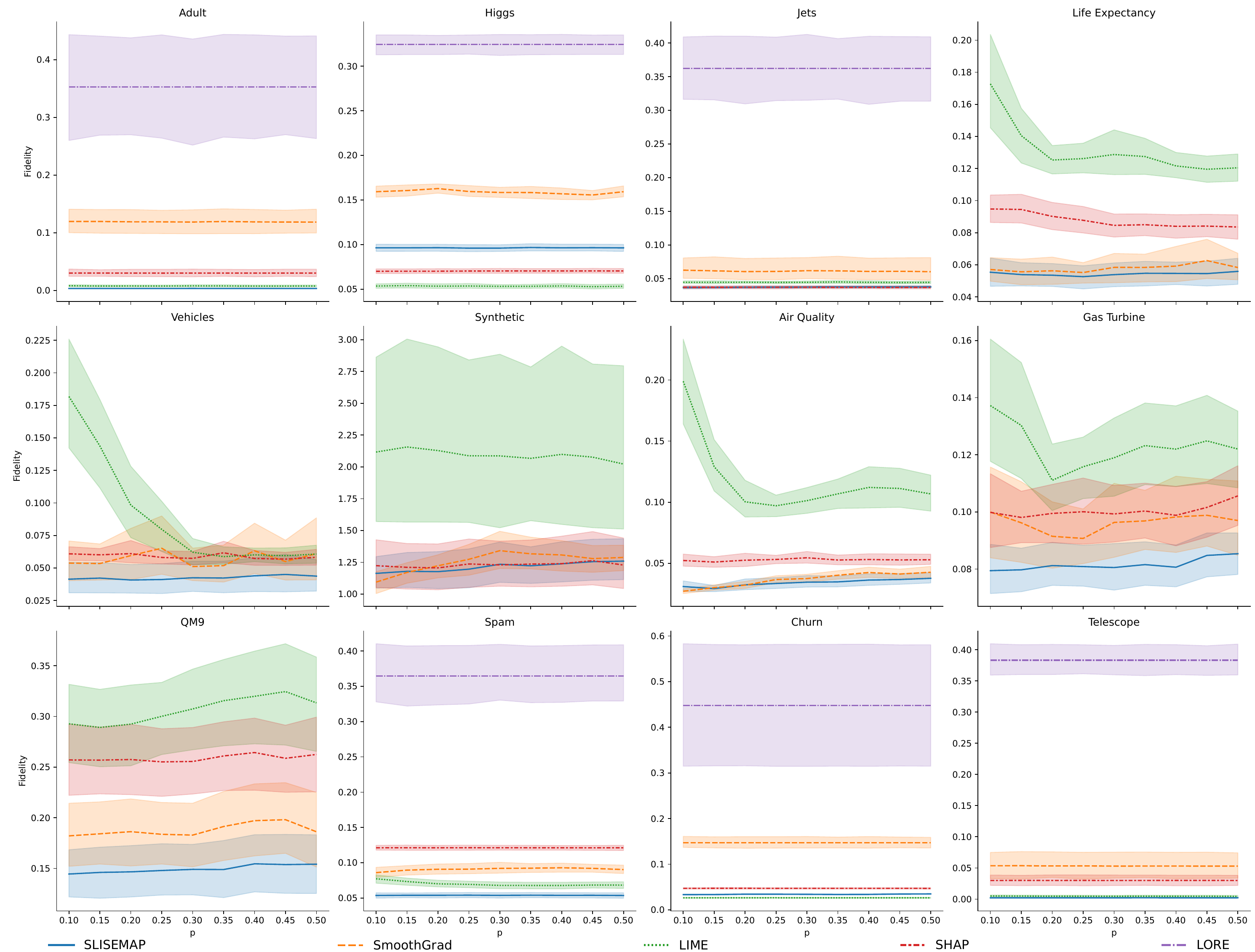}
    \caption{Test set fidelity as a function of error tolerance for the {\sc Balanced} algorithm.}
    \label{fig:coverage_p_full}
\end{figure}

\begin{figure}[ht]
    \centering
    \includegraphics[width=\linewidth]{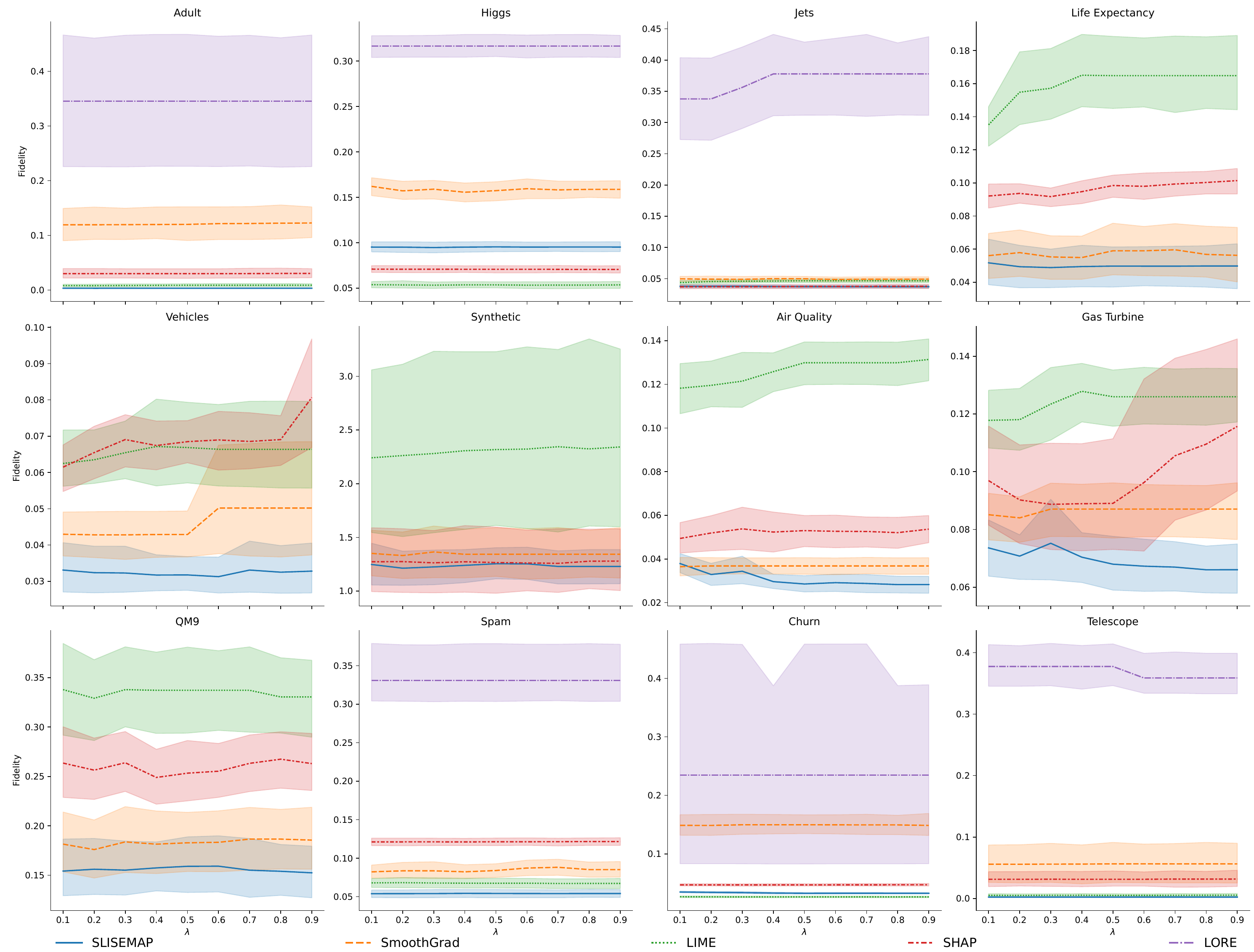}
    \caption{Test set fidelity as a function of trade-off parameter $\lambda$ for the {\sc Balanced} algorithm.}
    \label{fig:lambda_fidelity}
\end{figure}

\begin{figure}[ht]
    \centering
    \includegraphics[width=\linewidth]{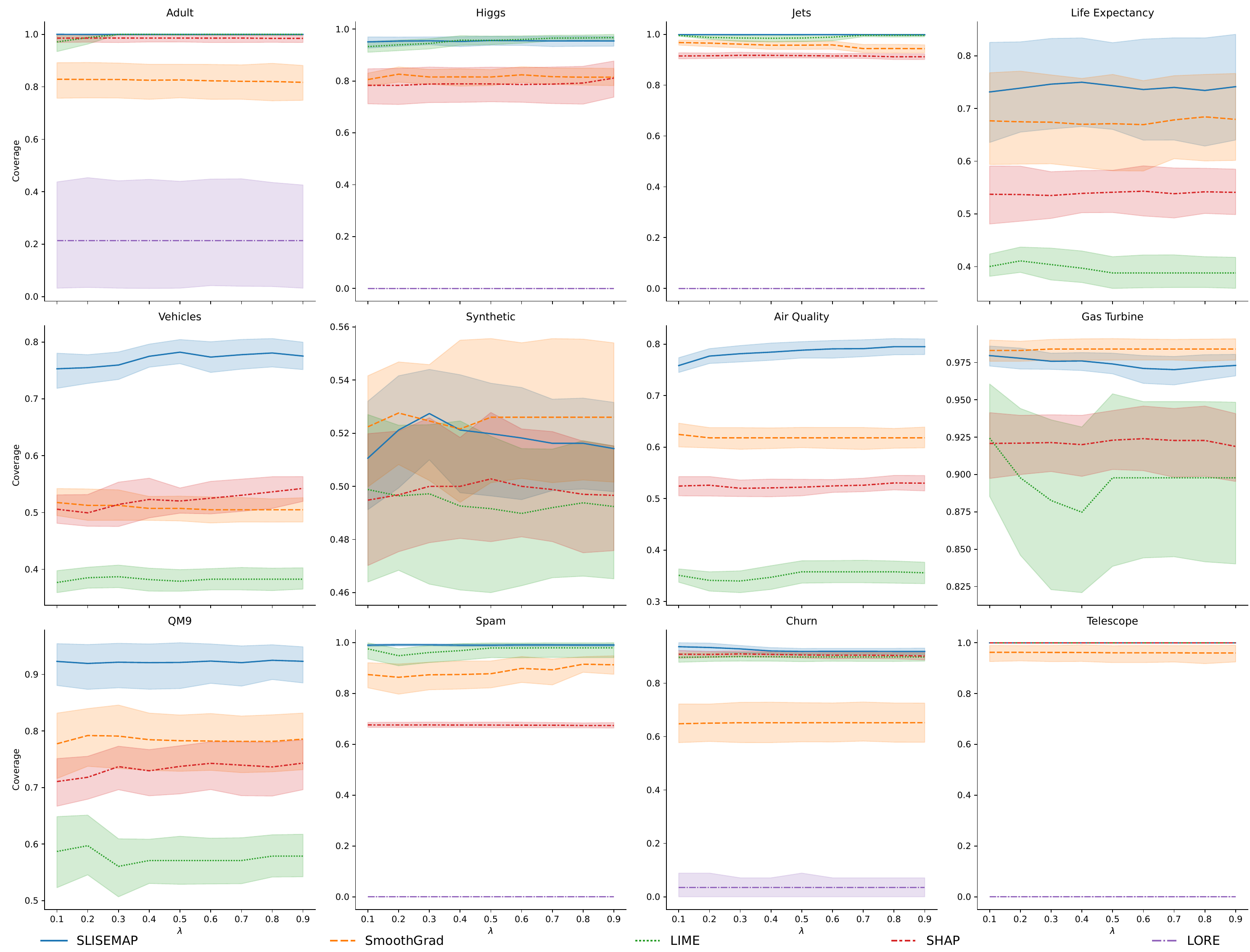}
    \caption{Coverage as a function of trade-off parameter $\lambda$ for the {\sc Balanced} algorithm.}
    \label{fig:lambda_coverage}
\end{figure}

As Fig. \ref{fig:coverage_p_full} shows, the {\sc Balanced} algorithm produces results with nearly identical fidelity across a wide range $\varepsilon$ at proxy set size $k=6$ and an initial set of $n=500$ local explanations.
The test fidelity of the proxy set for all tested reduction methods exhibits low variance when the error tolerance is set anywhere between the 10th and 50th percentiles of the closed-box training loss values.
Similarly, the effect of the trade-off parameter $\lambda$ is negligible at proxy set size $k=6$ (initial explanation set of $n=500$ explanations), as shown in Figures \ref{fig:lambda_fidelity} and \ref{fig:lambda_coverage}.
For simplicity, we suggest using $\lambda = 0.5$ and $\varepsilon = q(\bm{L}_{BB}, 0.3)$ as default arguments.
When not stated otherwise, these are also the $\lambda$ and $\varepsilon$ values used in other experiments in this paper.


\FloatBarrier  %
\clearpage  %

\section{Comparing {\sc ExplainReduce} to other model aggregation methods}
\label{a:global_comparison}

\begin{figure}
    \centering
    \includegraphics[width=\linewidth]{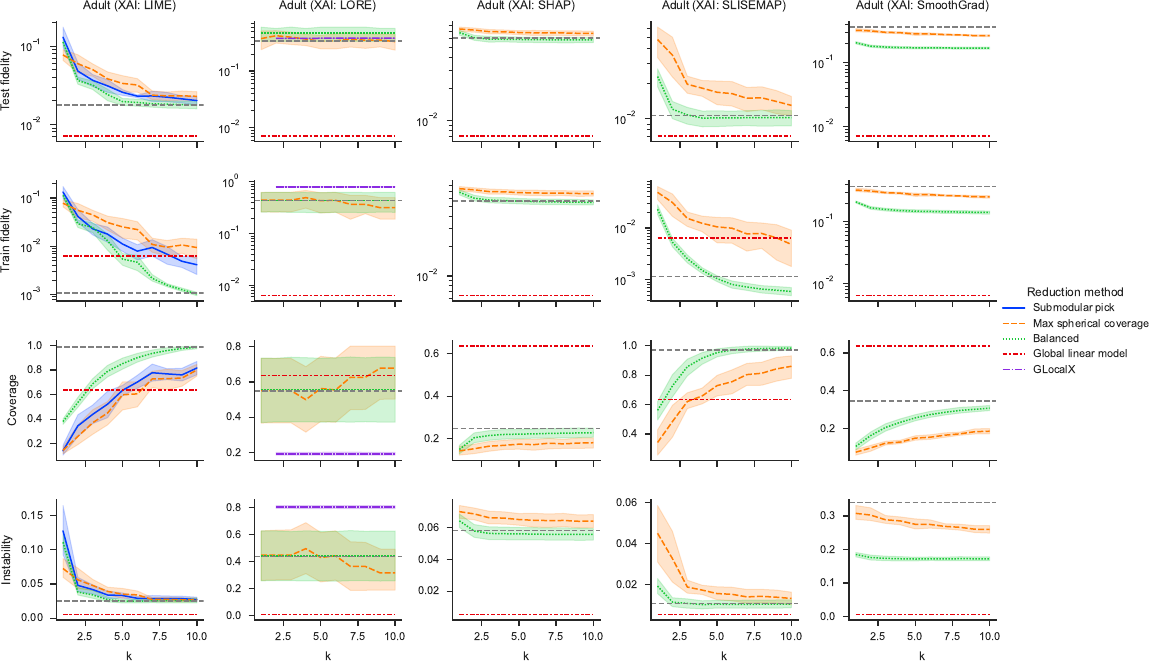}
    \caption{Adult dataset global comparison}
    \label{fig:gc_adult}
\end{figure}

\begin{figure}
    \centering
    \includegraphics[width=\linewidth]{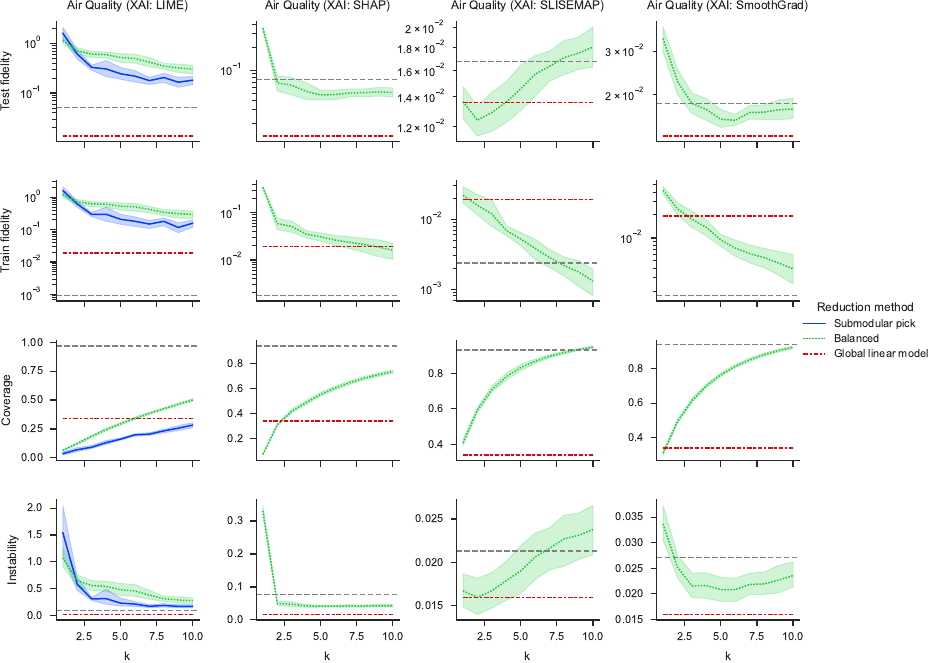}
    \caption{Air Quality dataset global comparison}
    \label{fig:gc_aq}
\end{figure}

\begin{figure}
    \centering
    \includegraphics[width=\linewidth]{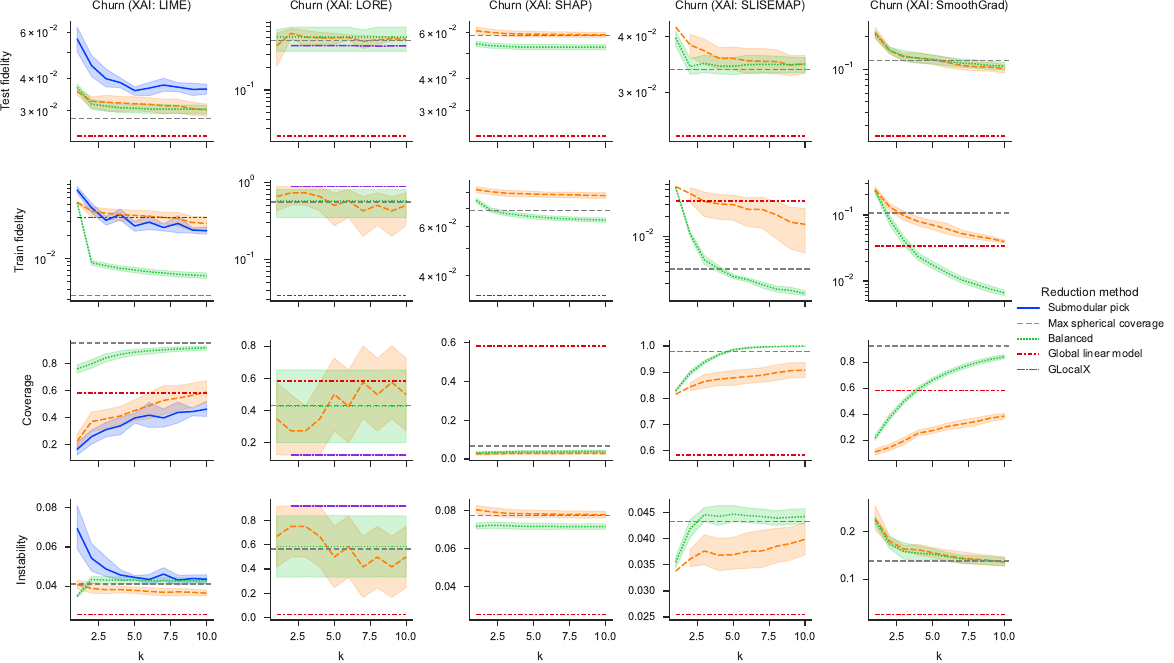}
    \caption{Churn dataset global comparison}
    \label{fig:gc_churn}
\end{figure}

\begin{figure}
    \centering
    \includegraphics[width=\linewidth]{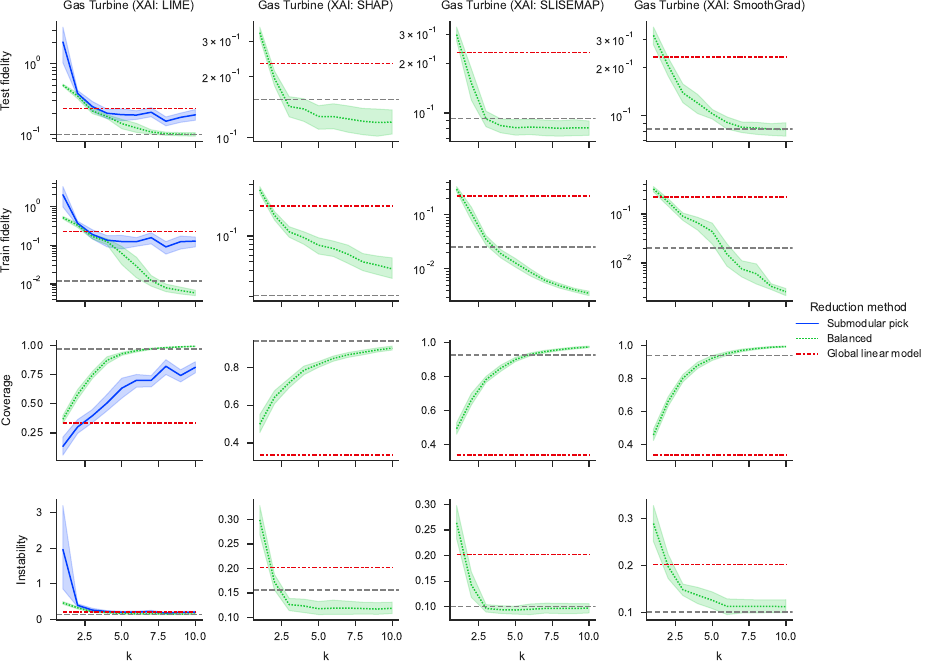}
    \caption{Gas Turbine dataset global comparison}
    \label{fig:gc_gt}
\end{figure}

\begin{figure}
    \centering
    \includegraphics[width=\linewidth]{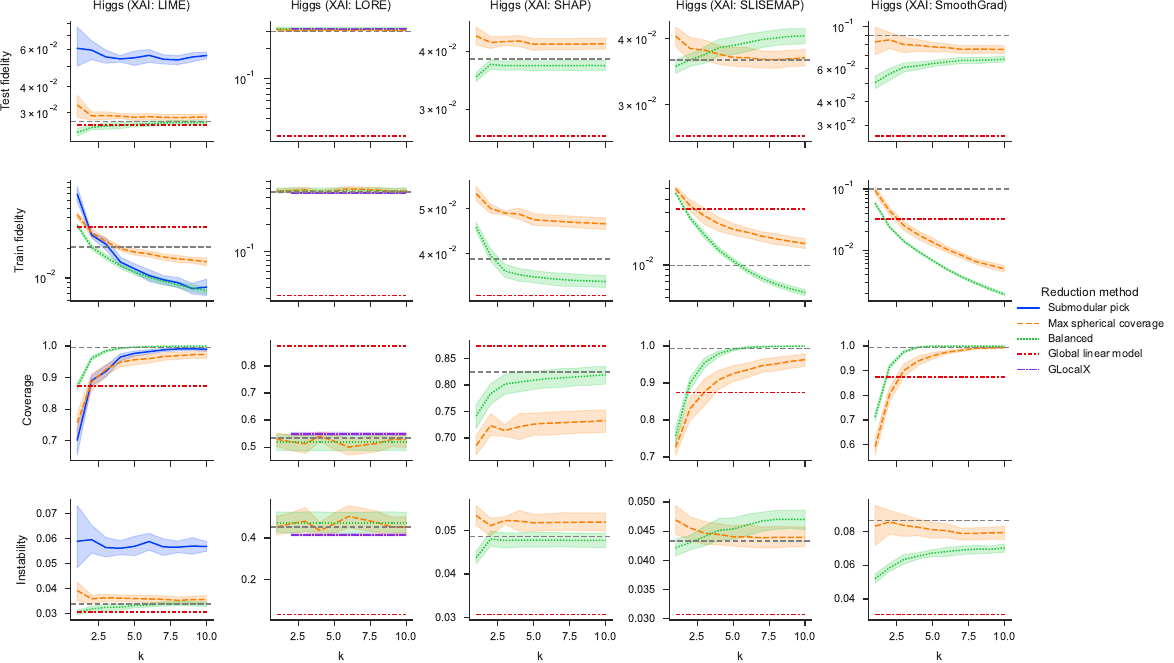}
    \caption{Higgs dataset global comparison}
    \label{fig:gc_higgs}
\end{figure}

\begin{figure}
    \centering
    \includegraphics[width=\linewidth]{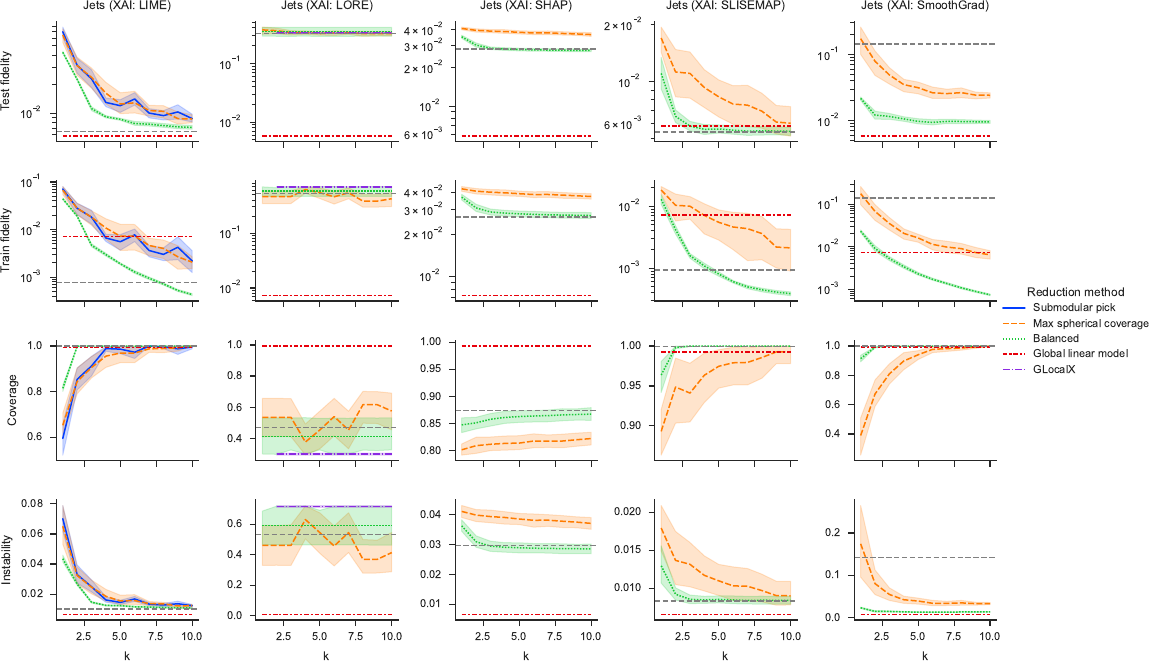}
    \caption{Jets dataset global comparison}
    \label{fig:gc_jets}
\end{figure}

\begin{figure}
    \centering
    \includegraphics[width=\linewidth]{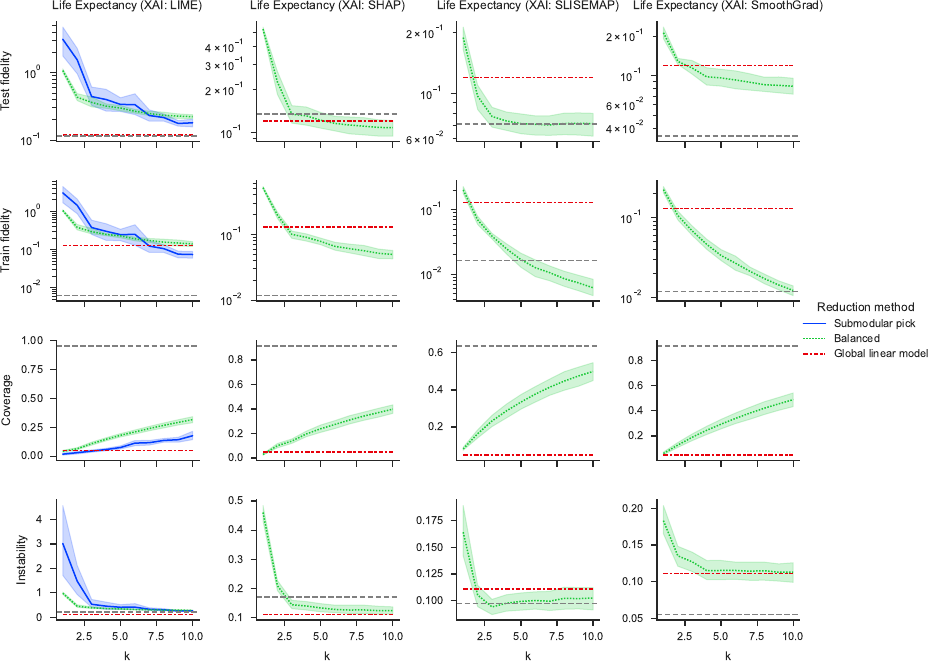}
    \caption{Life Expectancy dataset global comparison}
    \label{fig:gc_le}
\end{figure}

\begin{figure}
    \centering
    \includegraphics[width=\linewidth]{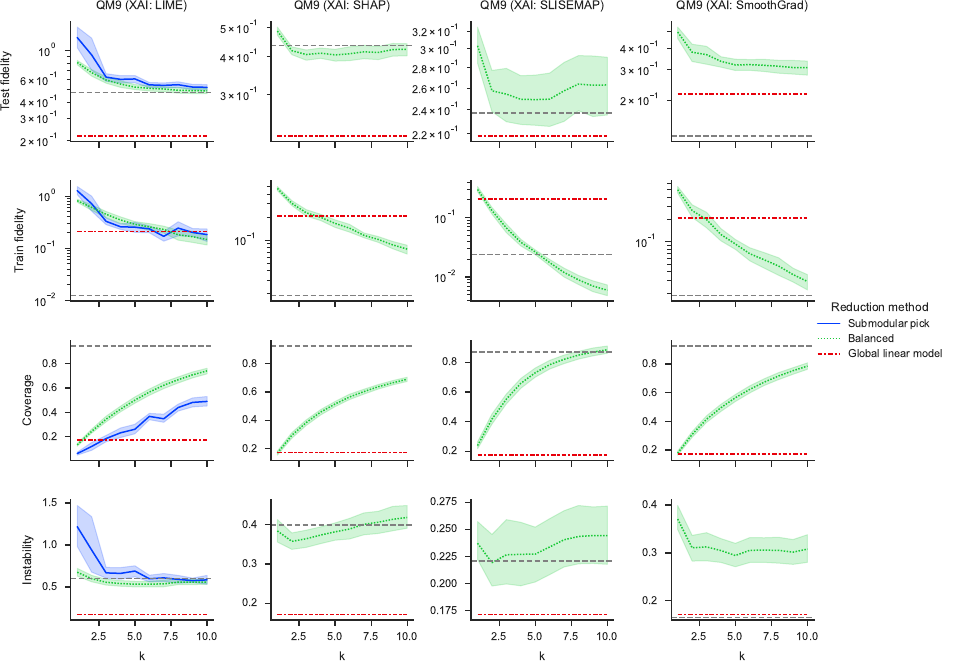}
    \caption{QM9 dataset global comparison}
    \label{fig:gc_qm9}
\end{figure}

\begin{figure}
    \centering
    \includegraphics[width=\linewidth]{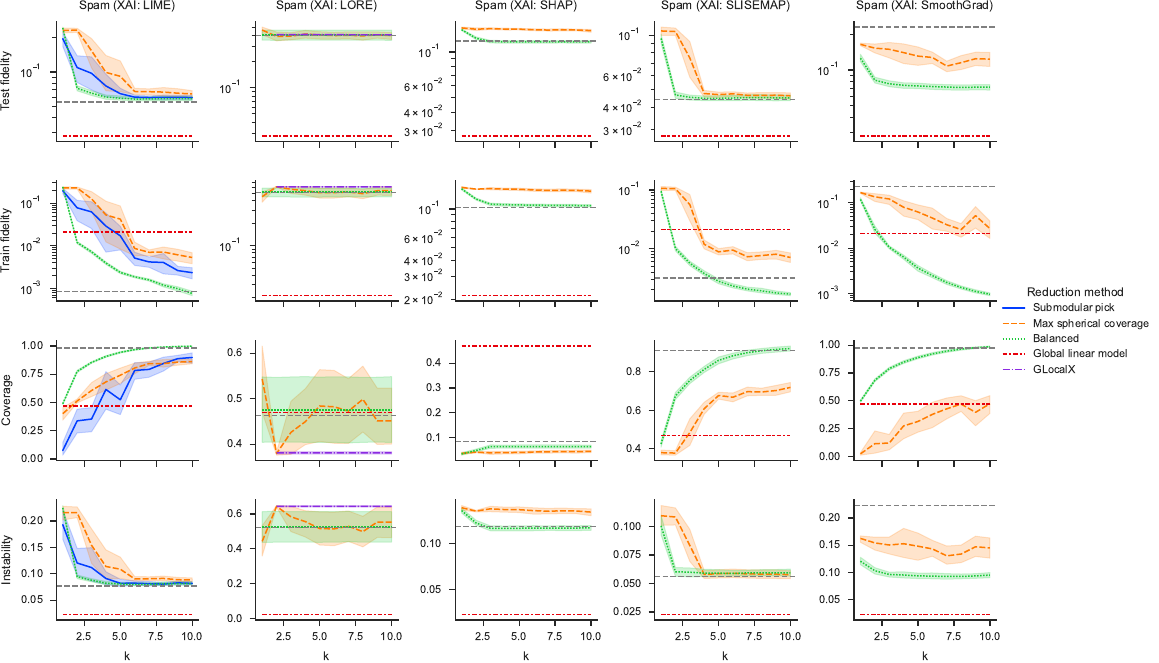}
    \caption{Spam dataset global comparison}
    \label{fig:gc_spam}
\end{figure}

\begin{figure}
    \centering
    \includegraphics[width=\linewidth]{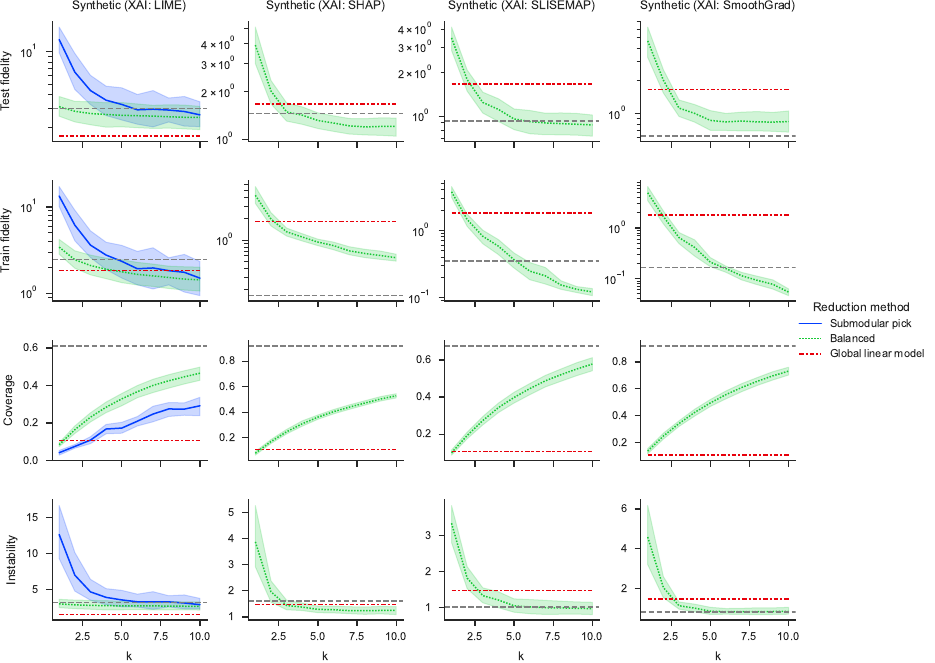}
    \caption{Synthetic dataset global comparison}
    \label{fig:gc_synth}
\end{figure}

\begin{figure}
    \centering
    \includegraphics[width=\linewidth]{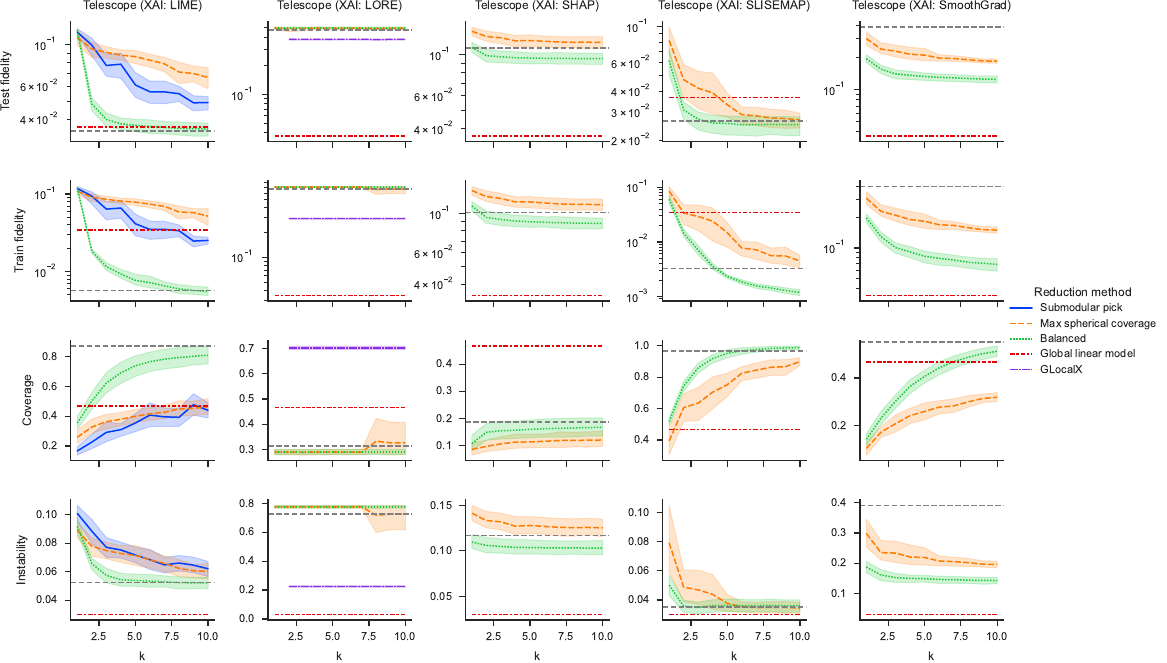}
    \caption{Telescope dataset global comparison}
    \label{fig:gc_telescope}
\end{figure}

\begin{figure}
    \centering
    \includegraphics[width=\linewidth]{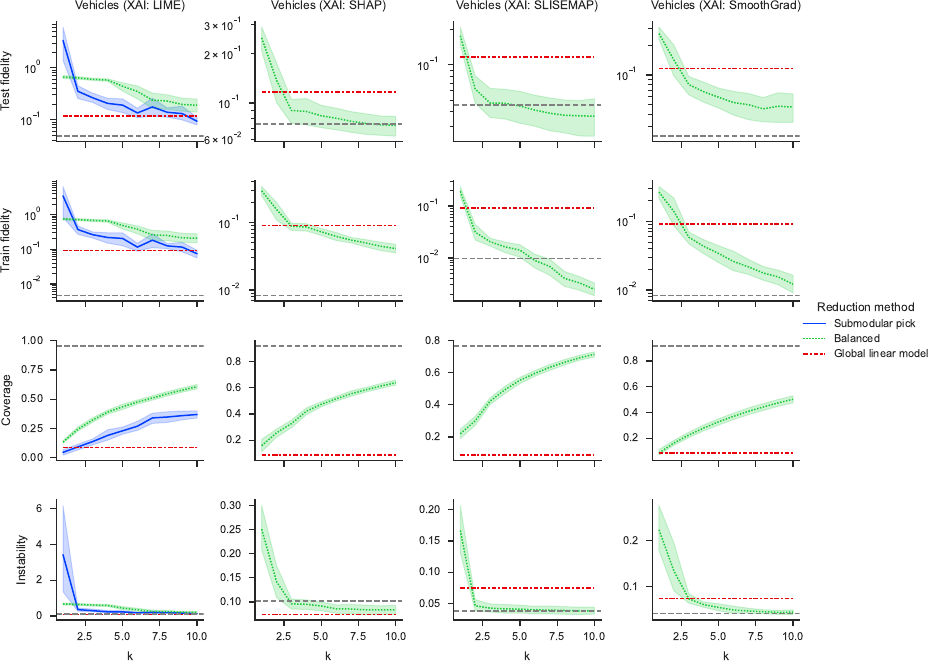}
    \caption{Vehicles dataset global comparison}
    \label{fig:gc_vehicles}
\end{figure}

Figures \ref{fig:gc_adult}-\ref{fig:gc_vehicles} show the test and train set fidelities, coverage and stability of {\sc ExplainReduce}, using the {\sc Balanced} algorithm, against submodular pick \citep{ribeiro2016}, GLocalX \citep{setzu2021glocalx}, the IP-based aggregation method by \cite{li2022optimal} and a single global linear explainer.
Except for GlocalX, which has a slight edge in the Telescope dataset (Fig. \ref{fig:gc_telescope}), our method consistently exceeds or at least matches the performance of other methods.

\FloatBarrier  %
\clearpage  %

\section{Performance of greedy algorithms}
\label{a:greedy}

In section \ref{sec:problem}, we discussed the performance of greedy approximation algorithms used to solve Problems \ref{prob:1}--\ref{prob:3}.
As we have seen in previous sections, the performance of the greedy and analytical coverage-optimising algorithms is almost equal as a function of the proxy set size and the initial local explanation set size.
As mentioned in the section above, solving for the exact loss-minimising proxy set is a computationally challenging problem which requires $\binom{m}{k}$ comparisons in the worst case.
Hence, the analysis in this section was limited to $n=m=100, k=5$ for computational tractability.
Table \ref{tab:greedy} shows how the greedy approximation algorithms perform comparably to the analytically optimal variants.
For the {\sc max coverage} algorithm, experimental approximation ratios are substantially better than the worst-case bound described in Sect. \ref{sec:problem}.
Furthermore, as we saw in the previous sections, while on the train set, the greedy heuristics do not reach the optimal solution fidelity-wise. However, on the test, the differences between the exact solution and the greedy heuristics vanish.
The {\sc Balanced} algorithm shows particularly poorer performance compared to the greedy {\sc min loss} algorithm, but it generalises better on the test set while achieving good coverage performance as well.
Overall, we can conclude that the greedy approximations offer a computationally feasible approach to generating proxy sets.

\begin{table}[t]
    \centering
    \begin{tabular}{l@{\hspace{3mm}} r@{\hspace{3mm}}r@{\hspace{3mm}}r@{\hspace{3mm}}r@{\hspace{3mm}}r@{\hspace{3mm}}r@{\hspace{3mm}}r}
\hline \\
\multicolumn{6}{|c|}{Coverage $\uparrow$}\\
\hline\\
\bfseries Dataset & \bfseries G. Max $c$ & \bfseries A. ratio & \bfseries Balanced & \bfseries A. ratio & \bfseries Max $c$ \\
\midrule
Air Quality & $0.82 \pm 0.18$ & $0.97 \pm 0.04$ & $0.8 \pm 0.17$ & $0.96 \pm 0.04$ & $0.84 \pm 0.18$ \\
Gas Turbine & $0.86 \pm 0.17$ & $0.99 \pm 0.01$ & $0.86 \pm 0.18$ & $0.99 \pm 0.03$ & $0.87 \pm 0.18$ \\
Jets & $0.86 \pm 0.17$ & $0.99 \pm 0.02$ & $0.86 \pm 0.17$ & $0.99 \pm 0.02$ & $0.87 \pm 0.18$ \\
QM9 & $0.78 \pm 0.17$ & $0.98 \pm 0.03$ & $0.78 \pm 0.17$ & $0.97 \pm 0.04$ & $0.8 \pm 0.18$ \\
\hline \\
\multicolumn{6}{|c|}{Train fidelity $\downarrow$}\\
\hline\\
\bfseries Dataset & \bfseries G. Min Loss & \bfseries A. ratio & \bfseries Balanced & \bfseries A. ratio & \bfseries Min Loss \\
\midrule
Air Quality & $0.36 \pm 0.35$ & $1.25 \pm 0.21$ & $0.47 \pm 0.43$ & $1.85 \pm 0.91$ & $0.32 \pm 0.35$ \\
Gas Turbine & $0.21 \pm 0.23$ & $2.08 \pm 0.9$ & $0.22 \pm 0.29$ & $1.75 \pm 0.81$ & $0.17 \pm 0.23$ \\
Jets & $0.02 \pm 0.02$ & $1.4 \pm 0.4$ & $0.02 \pm 0.03$ & $1.49 \pm 0.37$ & $0.01 \pm 0.02$ \\
QM9 & $0.53 \pm 0.46$ & $1.25 \pm 0.21$ & $0.59 \pm 0.47$ & $1.43 \pm 0.43$ & $0.48 \pm 0.47$ \\
\hline \\
\multicolumn{6}{|c|}{Test fidelity $\downarrow$}\\
\hline\\
\bfseries Dataset & \bfseries G. Min Loss & \bfseries A. ratio & \bfseries Balanced & \bfseries A. ratio & \bfseries Min Loss \\
\midrule
Air Quality & $0.4 \pm 0.29$ & $1.11 \pm 0.22$ & $0.47 \pm 0.37$ & $1.28 \pm 0.33$ & $0.37 \pm 0.28$ \\
Gas Turbine & $0.31 \pm 0.17$ & $1.2 \pm 0.22$ & $0.33 \pm 0.24$ & $1.17 \pm 0.16$ & $0.28 \pm 0.18$ \\
Jets & $0.02 \pm 0.01$ & $1.09 \pm 0.18$ & $0.02 \pm 0.02$ & $1.11 \pm 0.17$ & $0.02 \pm 0.01$ \\
QM9 & $1.19 \pm 0.13$ & $1.01 \pm 0.08$ & $1.14 \pm 0.13$ & $0.97 \pm 0.07$ & $1.18 \pm 0.11$ \\
\bottomrule
\end{tabular}

    \caption{Comparison of the training fidelity and coverage, alongside the observed approximation ratios (A. ratio), between the analytical and approximate optimisation algorithms when applied to $n=100$ explanations produced with {\sc lime} and with $k=5$ proxies. G. Max $c$ corresponds to the greedy {\sc max coverage} algorithm, while G. Min Loss and G. Min Loss (min $c$) correspond to the greedy {\sc min loss} and {\sc Balanced} algorithms, respectively.}
    \label{tab:greedy}
\end{table}

\FloatBarrier  %
\clearpage  %

\section{Formalisation of the reduction algorithms} \label{a:proofs}
\subsection{Max Coverage}
\begin{prob}[Max $k$-Coverage Variant]\label{prob:a1}
Given a dataset $\mathcal{D} = \{(\bm{x}_j, \bm{y}_j)\}_{j=1}^n$, a set of candidate models $\bm{G} = \{g_1, \dots, g_m\}$, and a loss threshold $\varepsilon > 0$, find a subset $\bm{S} \subseteq \bm{G}$ with $|\bm{S}| \leq k$ that maximizes the number of covered samples.
\end{prob}

\begin{definition}[Coverage Set]
For each model $g_i \in \bm{G}$, we define its coverage set $C_i \subseteq \mathcal{D}$ as the set of points for which the loss is within the threshold $\varepsilon$:
\begin{equation}
    C_i = \{ (\bm{x}_j, \bm{y}_j) \in \mathcal{D} \mid \ell(g_i(\bm{x}_j), \bm{y}_j) \leq \varepsilon \}
\end{equation}
\end{definition}

\begin{definition}[Objective Function]
The total coverage of a subset of models $\bm{S} \subseteq \bm{G}$ is given by the set function $f: 2^{\bm{G}} \to \mathbb{Z}_{\geq 0}$:
\begin{equation}
    f(\bm{S}) = \left| \bigcup_{g_i \in \bm{S}} C_i \right|
\end{equation}
\end{definition}

\begin{theorem}
The greedy algorithm for Problem \ref{prob:a1} yields a solution $\bm{S}_{greedy}$ such that:
\begin{equation}
    f(\bm{S}_{greedy}) \geq \left( 1 - \left( \frac{k-1}{k} \right)^k \right) f(\bm{S}^*) \geq \left( 1 - \frac{1}{e} \right) f(\bm{S}^*)
\end{equation}
where $S^*$ is the optimal subset of size $k$.
\end{theorem}

\begin{proof}
The problem is a direct mapping to the \textsc{Max $k$-Cover} problem. We observe the following properties of the objective function $f(\bm{S})$:
\begin{enumerate}
    \item \textbf{Monotonicity}: For any $\bm{A} \subseteq \bm{B} \subseteq \bm{G}$, $f(\bm{A}) \leq f(\bm{B})$, as the union of additional sets cannot decrease the cardinality of covered elements.
    \item \textbf{Submodularity}: For any $\bm{A} \subseteq \bm{B} \subseteq \bm{G}$ and $g \in \bm{G} \setminus B$, the marginal gain satisfies $f(\bm{A} \cup \{g\}) - f(\bm{A}) \geq f(\bm{B} \cup \{g\}) - f(\bm{B})$. This follows from the property of set unions where an element already covered by $\bm{A}$ is necessarily covered by $\bm{B}$.
\end{enumerate}
By the results established in \citet{nemhauser1978analysis} for the maximization of monotone submodular functions under cardinality constraints, the greedy selection strategy—which iteratively picks $g = \arg\max_{g \in \bm{G}} (f(\bm{S} \cup \{g\}) - f(\bm{S}))$—achieves the stated lower bound.
\end{proof}

\subsection{Min Loss}

\begin{prob}[Min Loss via Improvement Maximization] \label{prob:a2}
Given a dataset $\mathcal{D}$, find a subset $S \subseteq \bm{G}$ with $|S| = k$ that maximizes the total improvement function:
\begin{equation}
    f(\bm{S}) = \mathcal{L}_{base} - \mathcal{L}(\bm{S})
    \label{eq:emp_min_loss}
\end{equation}
where $\mathcal{L}(\bm{S}) = \frac{1}{n} \sum_{j=1}^n \min_{g_i \in S} \ell(g_i(\bm{x}_j), \bm{\hat{y}}_j)$ and $\mathcal{L}_{base}$ is the loss of the baseline configuration.
\end{prob}

\begin{lemma}[Supermodularity of Loss]
The function $\mathcal{L}: 2^{\bm{G}} \to \mathbb{R}_{\geq 0}$ defined in Eq.~\eqref{eq:emp_min_loss} is supermodular.
\end{lemma}

\begin{proof}
A function $\mathcal{L}$ is supermodular if for all $\bm{A} \subset \bm{B} \subset \bm{G}$ and $v \notin \bm{B}$:
\begin{equation}
    \mathcal{L}(\bm{A} \cup \{v\}) - \mathcal{L}(\bm{A}) \leq \mathcal{L}(\bm{B} \cup \{v\}) - \mathcal{L}(\bm{B}) \label{eq:supermod}
\end{equation}
Let $L_j(\bm{S}) = \min_{g_i \in S} \ell(g_i(\bm{x}_j), \bm{y}_j)$. Since $\mathcal{L}(\bm{S}) = \frac{1}{n} \sum_j L_j(\bm{S})$, and the sum of supermodular functions is supermodular, it suffices to prove the property for a single point $j$. Let $\ell_i = \ell(g_i(\bm{x}_j), \bm{y}_j)$. We evaluate the marginal change $\Delta(\bm{S}, v) = L_j(\bm{S} \cup \{v\}) - L_j(\bm{S})$:
\begin{equation}
    \Delta(\bm{S}, v) = \min(\min_{i \in \bm{S}} \ell_i, \ell_v) - \min_{i \in \bm{S}} \ell_i = \min(0, \ell_v - \min_{i \in \bm{S}} \ell_i)
\end{equation}
Since $\bm{A} \subset \bm{B}$, it follows that $\min_{i \in \bm{A}} \ell_i \geq \min_{i \in \bm{B}} \ell_i$. Let $M_{\bm{A}} = \min_{i \in \bm{A}} \ell_i$ and $M_{\bm{B}} = \min_{i \in \bm{B}} \ell_i$. Then $M_{\bm{A}} \geq M_{\bm{B}}$. The marginal changes are:
\begin{align*}
    \Delta(\bm{A}, v) &= \min(0, \ell_v - M_{\bm{A}}) \\
    \Delta(\bm{B}, v) &= \min(0, \ell_v - M_{\bm{B}})
\end{align*}
Since $M_{\bm{A}} \geq M_{\bm{B}}$, then $-M_{\bm{A}}\leq \min(0, \ell_v - M_{\bm{B}})$, satisfying \eqref{eq:supermod}.
\end{proof}

\begin{lemma}[Submodularity of Loss Reduction]
The improvement function $f(\bm{S})$ is a non-negative, monotone submodular function.
\label{lemma:loss_reduction_submodularity}
\end{lemma}

\begin{proof}
Let $h(\bm{S}) = -\mathcal{L}(\bm{S})$. We previously established that $\mathcal{L}(\bm{S})$ is supermodular; therefore, $h(\bm{S})$ is submodular \citep{mccormick2005submodular}.
\begin{enumerate}
    \item \textbf{Submodularity}: $f(\bm{S}) = \mathcal{L}_{base} + h(\bm{S})$. Since adding a constant does not change the second-order differences of a set function, $f(\bm{S})$ inherits the submodularity of $h(\bm{S})$.
    \item \textbf{Monotonicity}: For any $g \in \bm{G} \setminus \bm{S}$, the marginal gain is $\Delta f(g|\bm{S}) = f(\bm{S} \cup \{g\}) - f(\bm{S}) = \mathcal{L}(\bm{S}) - \mathcal{L}(\bm{S} \cup \{g\})$. Since $\min_{i \in S \cup \{g\}} \ell_i \leq \min_{i \in S} \ell_i$, it follows that $\mathcal{L}(\bm{S} \cup \{g\}) \leq \mathcal{L}(\bm{S})$, making the marginal gain $\Delta f(g|\bm{S}) \geq 0$.
    \item \textbf{Non-negativity}: Provided $\mathcal{L}_{base} \geq \mathcal{L}(\bm{S})$ for all $\bm{S}$ (which holds if $\mathcal{L}_{base}$ is the loss of the empty set or a single-model baseline), $f(\bm{S}) \geq 0$.
\end{enumerate}
\end{proof}

\begin{theorem}
The greedy algorithm (\textsc{reduce}) selecting $g = \arg\max \Delta \mathcal{L}(g|S)$ yields a solution $S_g$ such that:
\begin{equation}
    \mathcal{L}_{base} - \mathcal{L}(\bm{S}_g) \geq \left( 1 - \frac{1}{e} \right) (\mathcal{L}_{base} - \mathcal{L}(\bm{S}^*))
\end{equation}
where $S^*$ is the optimal subset of size $k$.
\end{theorem}

\begin{proof}
By Lemma \ref{lemma:loss_reduction_submodularity}, $f(\bm{S})$ satisfies the conditions of \citet{nemhauser1978analysis}. The greedy algorithm maximizes the marginal improvement at each step, ensuring the $(1-1/e)$ approximation ratio relative to the maximum possible improvement.
\end{proof}

\subsection{Formalization of Joint Utility Maximization}

\begin{prob}[Balanced Utility Maximization] \label{prob:a3}
Find a subset $\bm{S} \subseteq \bm{G}$ with $|\bm{S}| = k$ that maximizes the joint utility function:
\begin{equation}
    U(\bm{S}) = \lambda \cdot C(\bm{S}, \varepsilon) + (1 - \lambda) \cdot R(\bm{S})
\end{equation}
where $C(\bm{S}, \varepsilon)$ is the coverage function and $R(\bm{S}) = \frac{\mathcal{L}_{base} - \mathcal{L}(\bm{S})}{\mathcal{L}_{base}}$ is the normalized loss reduction.
\end{prob}

\begin{lemma}[Preservation of Submodularity]
The joint utility function $U(\bm{S})$ is monotone submodular for any $\lambda \in [0, 1]$.
\end{lemma}

\begin{proof}
Consider the two components of $U(\bm{S})$:
\begin{enumerate}
    \item \textbf{Coverage $C(\bm{S}, \varepsilon)$}: As established in Problem \ref{prob:a1}, this is a standard set-cover function, which is monotone and submodular.
    \item \textbf{Loss Reduction $R(\bm{S})$}: Let $\mathcal{L}(\bm{S})$ be the supermodular loss function from Problem \ref{prob:2}. Then $-\mathcal{L}(\bm{S})$ is submodular. Since $R(\bm{S}) = \frac{\mathcal{L}_{base}}{\mathcal{L}_{base}} - \frac{\mathcal{L}(\bm{S})}{\mathcal{L}_{base}} = 1 - \frac{1}{\mathcal{L}_{base}}\mathcal{L}(\bm{S})$, $R(\bm{S})$ is a linear transformation of a submodular function with a positive scaling of the negative term. Thus, $R(\bm{S})$ is submodular.
\end{enumerate}
Since $\lambda$ and $(1-\lambda)$ are non-negative, and the sum of submodular functions is submodular, $U(\bm{S})$ is submodular. Monotonicity follows because adding a model can only increase coverage and decrease (or maintain) loss.
\end{proof}

\begin{theorem}
The greedy algorithm $\bm{S}_g$ for Problem \ref{prob:3} satisfies:
\begin{equation}
    U(\bm{S}_g) \geq \left( 1 - \frac{1}{e} \right) U(\bm{S}^*)
\end{equation}
where $\bm{S}^*$ is the optimal subset of size $k$.
\end{theorem}

\begin{proof}
This is a direct application of the result by \citet{nemhauser1978analysis}. Since $U(\bm{S})$ is a monotone submodular function with $U(\emptyset) = 0$ (assuming $\mathcal{L}(\emptyset) = \mathcal{L}_{base}$), the greedy strategy of picking $g = \arg\max_{g \notin S} (U(\bm{S} \cup \{g\}) - U(\bm{S}))$ achieves the $(1-1/e)$ lower bound.
\end{proof}

\end{document}